%% file: main.tex
\documentclass[10pt,twocolumn,letterpaper]{article}

\usepackage[pagenumbers]{iccv} %

\input{packages}

\definecolor{iccvblue}{rgb}{0.21,0.49,0.74}
\usepackage[pagebackref,breaklinks,colorlinks,allcolors=iccvblue]{hyperref}

\def\paperID{6593} %
\def\confName{ICCV}
\def\confYear{2025}

\title{Beyond the Frame: Generating 360$^\circ$ Panoramic Videos from Perspective Videos}

\author{
  \begin{tabular}{ccccc}
    Rundong Luo\textsuperscript{1} & Matthew Wallingford\textsuperscript{2} & Ali Fahardi\textsuperscript{2}  & Noah Snavely\textsuperscript{1} & Wei-Chiu Ma\textsuperscript{1}
  \end{tabular} \\ [2ex]
  \textsuperscript{1}Cornell University \hspace{3mm}
  \textsuperscript{2}University of Washington
}

\input{macros}

\input{math_commands}

\newcommand{\embedTeaserVideo}{Embed teaser video}
\newcommand{\embedStabilizationVideo}{Embed stab video}
\newcommand{\embedRotationVideo}{Embed rotation video}

\newcommand{\embedPanoVideo}{Embed panodiffusion comparison video}

\begin{document}

\twocolumn[{%
\renewcommand\twocolumn[1][]{#1}%
\maketitle
\vspace{-12mm}
\captionsetup{type=figure}
\begin{center}
    \IfDefinedSwitch{\embedTeaserVideo}{\animategraphics[autoplay,loop,controls={play,stop}, width=\linewidth, trim=0 5mm 0 2mm, clip]{5}{figures-ICCV/teaser/v1/}{000}{044}
    }
    {\includegraphics[width=\linewidth, trim=0 5mm 0 2mm]{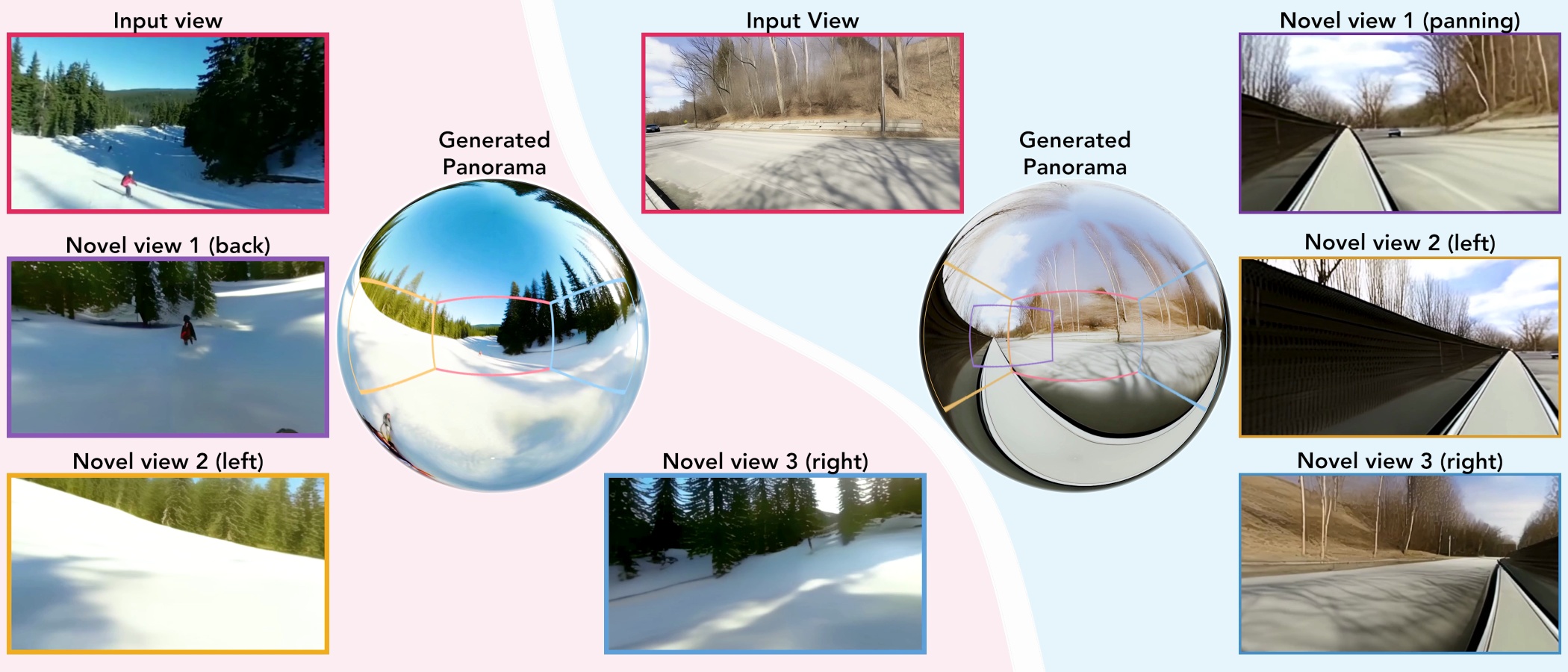}
    }
\end{center}
\vspace{-7mm}
\captionof{figure}{\textbf{360$^\circ$ videos generated by our model,~\modelname$^\dag$.} Starting from an input perspective video with arbitrary camera motion (\textcolor{MyRed}{red} box), \modelname~generates a full 360$^\circ$ panoramic video (visualized as environmental maps), where the \textcolor{MyRed}{red} box indicates the input view in the generated frame. The \textcolor{MyBlue}{blue}, \textcolor{MyOrange}{orange}, and \textcolor{MyPurple}{purple} boxes show sampled perspectives from the generated 360$^\circ$ video. \emph{Best viewed in Adobe Acrobat Reader for the \textbf{embedded videos}}.
}
\label{fig:teaser}
}
\vspace{3mm}
]

\input{sec/0_abstract_arxiv}
\input{sec/1_introduction}

\input{sec/2_related_works}

\input{sec/4_method}

\input{sec/5_experiments}

\input{sec/6_conclusion}

{
    \small
    \bibliographystyle{ieeenat_fullname}
    \bibliography{main}
}

\input{sec/7_supplementary_arxiv}

\end{document}

%% file: packages.tex
\usepackage{amsmath,amsfonts,amsthm,amssymb}
\usepackage{mathtools}
\usepackage{bm}
\usepackage{nicefrac}
\usepackage{microtype}
\usepackage{animate}

\usepackage{xcolor}
\usepackage{epsfig}
\usepackage{graphicx}

\usepackage{adjustbox}
\usepackage{array}
\usepackage{booktabs}
\usepackage{colortbl}
\usepackage{wrapfig}
\usepackage{hhline}
\usepackage{multirow}
\usepackage[size=small]{caption}

\usepackage{changepage}
\usepackage{extramarks}
\usepackage{fancyhdr}
\usepackage{lastpage}
\usepackage{setspace}
\usepackage{soul}
\usepackage{xspace}
\usepackage{adjustbox}

\usepackage{algpseudocode}
\usepackage{algorithmicx}
\usepackage[ruled]{algorithm2e}
\usepackage{enumitem}  %
\usepackage{makecell}
\usepackage{pifont}
\usepackage{fontawesome}
\usepackage{ifthen}
\usepackage{etoolbox}

%% file: macros.tex
\newcolumntype{L}[1]{>{\raggedright\let\newline\\\arraybackslash\hspace{0pt}}m{#1}}
\newcolumntype{C}[1]{>{\centering\let\newline\\\arraybackslash\hspace{0pt}}m{#1}}
\newcolumntype{R}[1]{>{\raggedleft\let\newline\\\arraybackslash\hspace{0pt}}m{#1}}

\newcommand{\ignore}[1]{}

\newcommand{\projectpageurl}{https://red-fairy.github.io/argus/}
\newcommand{\projectpagehref}{\href{https://red-fairy.github.io/argus/}{project page}}

\DeclareMathAlphabet{\mathbfit}{OML}{cmm}{b}{it}

\makeatletter
\DeclareRobustCommand\onedot{\futurelet\@let@token\@onedot}
\def\@onedot{\ifx\@let@token.\else.\null\fi\xspace}

\def\modelname{Argus}
\makeatother

\definecolor{MyDarkBlue}{rgb}{0,0.08,1}
\definecolor{MyAqua}{rgb}{0,0.7,0.7}
\definecolor{MyDarkGreen}{rgb}{0.02,0.6,0.02}
\definecolor{MyDarkOrange}{rgb}{0.40,0.2,0.02}
\definecolor{MyPurple}{HTML}{7338a0}
\definecolor{MyPink}{RGB}{250, 147, 240}
\definecolor{MyRed}{RGB}{227, 44, 94}
\definecolor{MyBlue}{RGB}{69, 144, 201}
\definecolor{MyOrange}{HTML}{d6970e}
\definecolor{MyPink}{HTML}{D86ECC}
\definecolor{MyGold}{rgb}{0.75,0.6,0.12}
\definecolor{MyDarkgray}{rgb}{0.66, 0.66, 0.66}

\newcommand{\myparagraph}[1]{\vspace{0.01mm}\noindent\textbf{#1}}

\newcommand{\IfDefinedSwitch}[3]{%
  \ifdefined#1
    #2 %
  \else
    #3 %
  \fi
}

%% file: math_commands.tex
\usepackage{amsmath,amsfonts,bm}

\def\eqref#1{equation~\ref{#1}}

\def\1{\bm{1}}

\DeclareMathAlphabet{\mathsfit}{\encodingdefault}{\sfdefault}{m}{sl}
\SetMathAlphabet{\mathsfit}{bold}{\encodingdefault}{\sfdefault}{bx}{n}

%% file: sec/0_abstract_arxiv.tex
\begin{abstract}
360$^\circ$ videos have emerged as a promising medium to represent our dynamic visual world. Compared to the ``tunnel vision'' of standard cameras, their borderless field of view offers a more complete perspective of our surroundings. While existing video models excel at producing standard videos, their ability to generate full panoramic videos remains elusive. 
In this paper, we investigate the task of video-to-360$^\circ$ generation: given a perspective video as input, our goal is to generate a full panoramic video that is consistent with the original video. 
Unlike conventional video generation tasks, the output's field of view is significantly larger, and the model is required to have a deep understanding of both the spatial layout of the scene and the dynamics of objects to maintain spatio-temporal consistency. 
To address these challenges, we first leverage the abundant 360$^\circ$ videos available online and develop a high-quality data filtering pipeline to curate pairwise training data. 
We then carefully design a series of geometry- and motion-aware operations to facilitate the learning process and improve the quality of 360$^\circ$ video generation. 
\footnotetext{$^\dag$\modelname~is named after a figure in Greek mythology with many eyes, symbolizing the ability to observe from multiple perspectives.}
Experimental results demonstrate that our model can generate realistic and coherent 360$^\circ$ videos from in-the-wild perspective video. In addition, we showcase its potential applications, including video stabilization, camera viewpoint control, and interactive visual question answering. View more high-resolution video results~\href{\projectpageurl}{here}\footnotemark.\footnotetext{This file contains embedded videos best viewed in Adobe Acrobat Reader. High-resolution results are available on our~\projectpagehref.}

\end{abstract}

%% file: sec/1_introduction.tex
\section{Introduction}
\label{sec:intro}
Remarkable advances in video generation have led to impressive capabilities, driven in part by large-scale video data from the web~\cite{ho2022video, blattmann2023stable, singer2023make, ho2022imagen, blattmann2023align}. Models can now produce high-quality video clips based on an input image, allowing us to step into the world behind the pixels. While these models achieve impressive fidelity, they still provide us only a narrow slice of the four-dimensional scene. Unlike the real world where we can freely look around and observe events as they unfold, current video models are restricted to a narrow, fixed perspective. Expanding video to the 360$^\circ$ medium, which more faithfully captures the visual world, enables better understanding of spatial layout and scene dynamics while improving spatio-temporal coherence. For example, standard video models commonly suffer from spatio-temporal inconsistency where content changes when looking back at previously observed parts of the scene. However, we find that generating 360$^\circ$ videos naturally resolves this problem as the entire scene is consistently visible.

To this end, we study the task of video-to-360$^\circ$ generation, aiming to generate a complete 360$^\circ$ video of a dynamic scene from a single-view perspective video. 
This task is difficult as it poses the following challenges: the input video only offers a narrow range of viewpoints, while the model must comprehend both the spatial layout of the scene and the dynamics of objects, then extrapolate to the entire scene. As illustrated in Figure~\ref{fig:teaser}, when the model observes a vehicle entering and then existing the frame (the \textcolor{MyRed}{red} box), it must infer both the vehicle's previous and future trajectories and the progression of the surrounding scene. This prediction requires deep understanding of real-world constraints---for instance, that roads typically extend in a straight line, and vehicles maintain their lane at a constant pace.

One straightforward approach would be expanding the input video using existing video outpainting models~\cite{be-your-outpainter, chen2024follow, dehan2022complete, fan2023hierarchical}. However, as we will show in Section~\ref{sec:exp}, their generation quality degrades drastically as we extend further from the input viewpoint. This issue arises because current models are trained on videos with narrow field-of-view, which prevents them from learning complete scene dynamics.

To overcome these challenges, we leverage the relatively untapped data source of 360° videos. The growing popularity of 360° cameras has created a wealth of panoramic content spanning sports, travel, and everyday activities—providing valuable insights into how scenes and actions naturally unfold in our world. 
We formulate this task as a video outpainting problem from dynamic masks. Given a perspective video, our approach first estimates camera poses for each frame and projects them onto equirectangular maps within a shared coordinate system. We then condition a diffusion-based generation process on these maps and the input video. To facilitate model training, we propose three key techniques: camera motion simulation that models perspective video trajectories from 360° video, view-based frame alignment to ensure a fixed viewpoint in the generated panorama, and blended decoding to maintain boundary coherence. Our model, \modelname, is the first to generate realistic and coherent 360° videos from standard perspective inputs.

Experimental results demonstrate that \modelname~outperforms existing methods in spatial coherence and visual quality. Our approach maintains consistency between the input and the generated content while producing realistic panoramic videos. The model generalizes effectively to various data sources, including online clips, self-recorded videos with complex dynamics, and model generated videos. Furthermore, \modelname~opens possibilities for several downstream applications, including video stabilization, camera viewpoint control, dynamic environmental mapping, and interactive visual question answering.

%% file: sec/2_related_works.tex
\section{Related Works}
\myparagraph{Video Generation.} Video generation aims to create high-quality, temporally consistent videos from multimodal inputs. Researchers have explored various architectures, including RNNs~\cite{babaeizadeh2017stochastic, castrejon2019improved, wang2017predrnn, denton2018stochastic}, normalizing flows~\cite{blattmann2021ipoke, dorkenwald2021stochastic}, GANs~\cite{MoCoGAN, vondrick2016generating, gupta2022rv, luc2020transformation, tian2021good}, and transformers~\cite{yan2021videogpt, TATS, wu2022nuwa, wu2021godiva}. However, these approaches suffer from resolution limitations and poor generalization, as they primarily train on small datasets designed for discriminative tasks. The recent success of diffusion models~\cite{rombach2022high, cascaded-diffusion} and access to larger, high-quality datasets have accelerated progress in video generation. While these approaches~\cite{ho2022video, blattmann2023stable, singer2023make, ho2022imagen, yang2024cogvideox} produce remarkably realistic videos from text or image prompts, they remain constrained to narrow field-of-view outputs, preventing the generation of full 360° panoramic experiences.

\myparagraph{Video Outpainting.} While diffusion-based image outpainting from arbitrary mask regions has achieved satisfactory results by mask conditioning~\cite{saharia2022palette, rombach2022high} or inference process modifications~\cite{lugmayr2022repaint, corneanu2024latentpaint}, video outpainting is limited to rectangular frame extensions~\cite{be-your-outpainter, dehan2022complete, chen2024follow, fan2023hierarchical}, constraining its application in panoramic content generation. Recently, VidPanos~\cite{ma2024vidpanos} introduced a method for synthesizing video panoramas from panning footage, but it focuses on dynamics within the observed regions and cannot extrapolate beyond initial viewpoints. Our approach overcomes these limitations by enabling flexible outpainting across dynamic, non-linear regions within a complete 360$^{\circ}$ panorama, generating immersive 360$^{\circ}$ scenes from single-view video inputs. This advancement expands video outpainting capabilities, enabling the generation of content that captures the full spatial and temporal dynamics of environments.

\myparagraph{360° Panorama Generation.} Generating 360° panoramic content presents unique challenges due to nonlinear distortions in equirectangular projections. These distortions warp objects and spatial layouts, complicating geometric appearance and creating boundary discontinuities. While current 360° image panorama generation methods~\cite{akimoto2019360, oh2022bips, yuan2024camfreediff, wu2023panodiffusion, tang2023mvdiffusion, kalischek2025cubediff} produce satisfactory results, they struggle with video panoramas where temporal coherence and spatial consistency are crucial. For video panorama generation, Wang et al.~\cite{360dvd} proposed a text-to-360° video generation framework, emphasizing text alignment rather than video-to-panorama transformation. Most relevant to our work is~\cite{tan2024imagine360}, where Tan et al. independently developed a video-to-360° framework based on AnimateDiff~\cite{guo2023animatediff}. However, their approach assumes pitch-only camera movements, uses limited training data, and confines evaluation to model-generated, subject-less or subject-centered scenes with minimal camera movement. We address these problems through geometry- and motion-aware modules and larger-scale training data. Our method generates realistic 360° panoramic videos from perspective inputs, outperforming existing approaches.

%% file: sec/4_method.tex
\section{Video to 360°}
\label{sec:method}
Given a standard perspective video as input, our goal is to extrapolate beyond its limited field of view to generate a corresponding 360$^\circ$ panoramic video. 
The generated panorama must maintain both content consistency and temporal dynamics that align with the input frames.

Since the problem is heavily under-constrained, we propose to capitalize on a relatively untapped data source -- 360$^\circ$ videos -- to learn the priors.
We start with the 360-1M dataset~\cite{wallingford2024imagine}, which consist of approximately 1 million videos of varying quality, and systematically filter down to 283,863 video clips (see the supp. material for details). 
Then, we build upon a diffusion-based image-to-video architecture~\cite{rombach2022high, karras2022elucidating, blattmann2023stable} and introduce a series of geometry- and motion-aware design tailored for video-to-360$^\circ$ generation (\emph{e.g.},  camera motion simulation, view-based frame alignment, etc). As we will show in Section~\ref{sec:exp}, these modifications are crucial for generating realistic panoramic videos.

\subsection{Video-Conditioned 360° Diffusion}
\label{sec:diffusion}
Our goal is to learn a diffusion mapping between an input perspective video $X_{\text{pers}} \in \mathbb{R}^{T \times 3 \times H \times W}$ and an output 360$^\circ$ panoramic video $Y_{\text{equi}} \in \mathbb{R}^{T \times 3 \times H' \times W'}$. 
We represent 360$^\circ$ video frames as equirectangular images and denote the number of frames by $T$.
Following Latent Diffusion Models \cite{rombach2022high, karras2022elucidating, blattmann2023stable}, our model consists of an encoder $\mathcal{E}$, a decoder $\mathcal{D}$, an image feature extractor $\mathcal{F}$, and a denoising U-Net $f_\theta$, with $f_\theta$ as the only learnable component. We adopt the temporal VAE from Stable Video Diffusion~\cite{rombach2022high} as our encoder and decoder, while the feature extractor is CLIP~\cite{radford2021learning}.

Since diffusion models require the input and the output to have the same dimensionality, we first convert the input perspective video $X_{\text{pers}}$ into an equirectangular format $X_{\text{equi}}$, matching the dimensions of the output $Y_{\text{equi}}$. 
The unmapped areas are set to black .
Next, we encode both equirectangular videos, $X_{\text{equi}}$ and $Y_{\text{equi}}$, to continuous latents, $\mathbf{x}_\text{equi} = \mathcal{E}(X_{\text{equi}})$ and $\mathbf{y}_\text{equi} = \mathcal{E}(Y_{\text{equi}})$. Finally, we add time-dependent noise to $\mathbf{y}_\text{equi}$ to produce $\mathbf{y}_{\text{equi},t}$, concatenate it with a noise-augmented~\cite{cascaded-diffusion} version of $\mathbf{x}_{\text{equi}}$, and feed this combination into the denoising network $f_\theta$ to estimate the injected noise. The network $f_\theta$ is conditioned on the timestamp $t$ and the image feature sequence $\mathcal{F}(X_{\text{pers}})$ through cross-attention~\cite{rombach2022high}. 
In practice, projecting from perspective to equirectangular format requires prior knowledge of the camera's field of view and poses. While this information is known during training (determined when extracting perspective frames from 360$^\circ$ videos), it is unknown during inference. In Section \ref{sec:model-inference}, we will describe how we address this challenge.

\begin{figure}[t!]
    \centering
    \includegraphics[width=\linewidth]{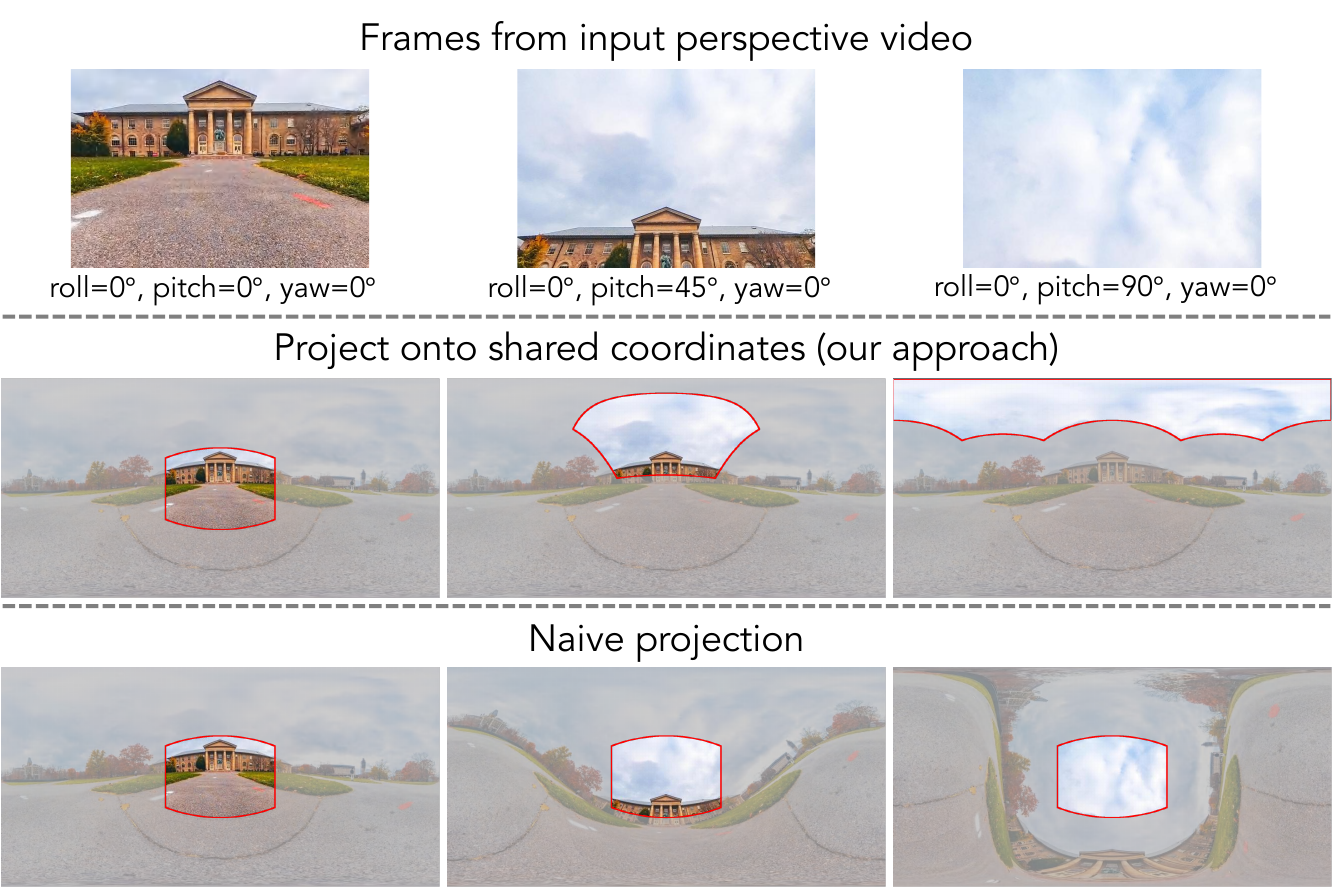}
    \vspace{-5.5mm}
    \caption{\textbf{View-based frame alignment.} Given input perspective video frames (first row), we project them onto shared coordinates to ensure a consistent viewing direction (second row). Without alignment, placing all video frames at the center (third row) forces the model to learn varying scene arrangements (\textit{e.g.}, the sky appearing at different heights), complicating the learning process.}
    \label{fig:projection}
    \vspace{-3mm}
\end{figure}

\begin{figure}[t!]
    \centering
    \includegraphics[width=\linewidth]{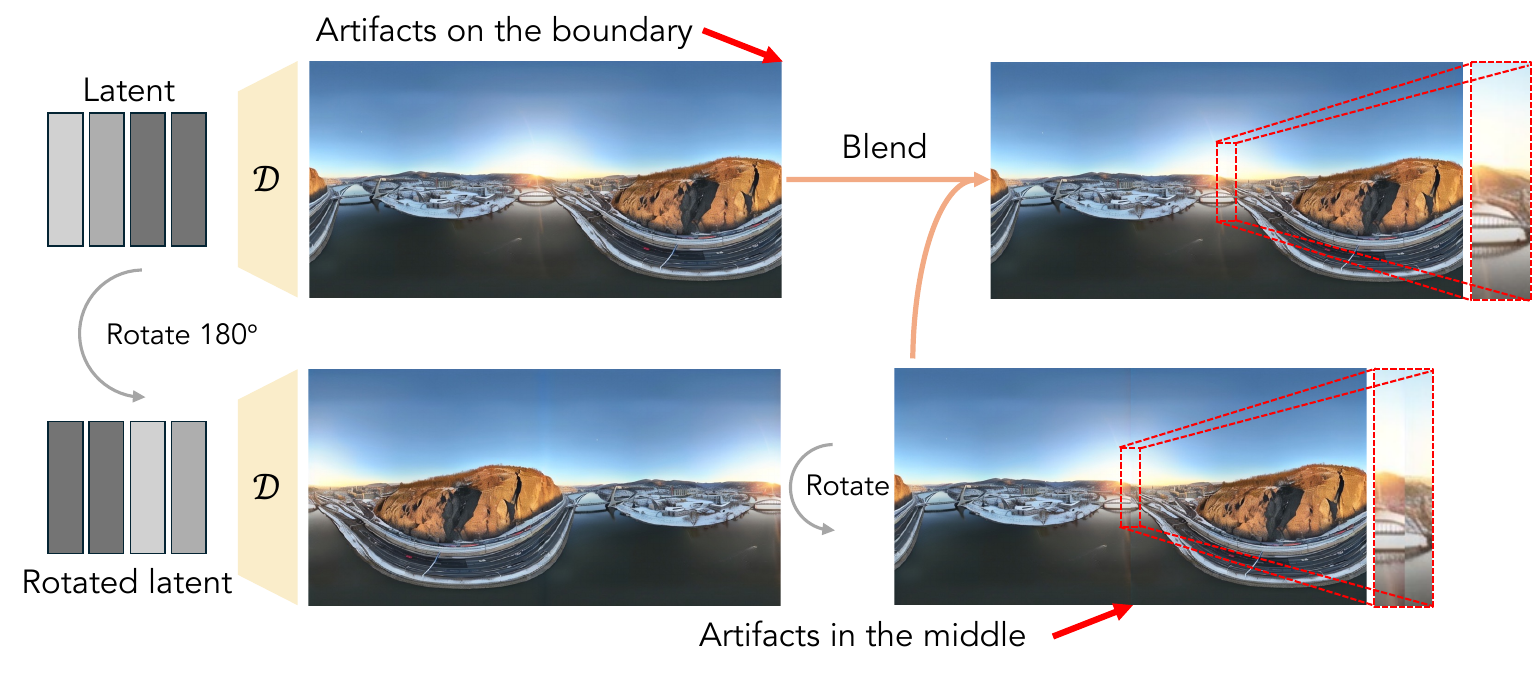}
    \vspace{-7.5mm}
    \caption{\textbf{Blended decoding.} We blend the video decoded from the original and 180$^\circ$-rotated latents to ensure boundary consistency. Zoom in to see the artifacts on the bottom-right image.}
    \label{fig:blend-decoding}
    \vspace{-2.5mm}
\end{figure}

\begin{table*}[t!]
\centering
\footnotesize
    \scalebox{0.89}{
        \begin{tabular}{lccccccccccccc}
        \toprule \vspace{0.2mm}
         \multirow{2}{*}{Method} & \multicolumn{6}{c}{Real camera trajectory} & \multicolumn{6}{c}{Simulated camera trajectory} & \multicolumn{1}{c}{Geometry} \\
         \cmidrule(lr){2-7}\cmidrule(lr){8-13}\cmidrule(lr){14-14}
         & PSNR$\uparrow$ & LPIPS$\downarrow$ & FVD$\downarrow$ & Imag.$\uparrow$ & Aes.$\uparrow$ & Motion$\uparrow$ & PSNR$\uparrow$ & LPIPS$\downarrow$ & FVD$\downarrow$ & Imag.$\uparrow$ & Aes.$\uparrow$ & Motion$\uparrow$ & Line cons.$\uparrow$ \\
        \midrule
            PanoDiffusion~\cite{wu2023panodiffusion} & 16.44 &	0.4138 & 2649.0 & \textbf{0.5055} & 0.4486 & 0.9426 & 15.28 & 0.4469 & 2622.3 & \textbf{0.4986} & 0.4533 & 0.9384 & 0.6504 \\
            \modelname~(ours) & \textbf{21.83} & \textbf{0.2409} & \textbf{1228.6} & 0.4939 & \textbf{0.4828} & \textbf{0.9802} & \textbf{21.50} & \textbf{0.2602} & \textbf{1100.1} & 0.4812 & \textbf{0.4784} & \textbf{0.9805} & \textbf{0.8506} \\
        \bottomrule
        \end{tabular}
    }
    \vspace{-2mm}
    \caption{\textbf{Quantitative results for video-to-360$^\circ$ generation.} We finetune PanoDiffusion~\cite{wu2023panodiffusion} on 360$^\circ$ video frames for fair comparison. \emph{Imag.}, \emph{Aes.}, and \emph{Motion} stands for the Imaging Quality, Aesthetic Quality, and Motion Smoothness metrics from VBench~\cite{vbench}. \emph{Line cons.} stands for our proposed \emph{line consistency} metric. Simulated trajectories are generated by our camera motion simulation technique, and real-world trajectories are extracted from in-the-wild videos through calibration.}
\label{tab:comparison-panodiffusion}
\end{table*}

\begin{figure*}[t]
    \begin{minipage}[t]{0.2328\textwidth}
        \centering
        \small Input Video \\ [0.05em]
        \IfDefinedSwitch{\embedPanoVideo}{
        \animategraphics[autoplay,loop,width=0.29\linewidth, trim={0 0 0 0}, clip]{5}{figures-ICCV/comparison-pano/27/input/}{000}{024} \\ [0.09em]
        \animategraphics[autoplay,loop,width=\linewidth, trim={0 0 0 0}, clip]{5}{figures-ICCV/comparison-pano/44/input/}{000}{024} 
        }{
        \includegraphics[width=0.29\linewidth]{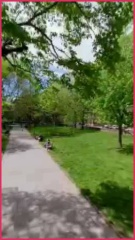} \\ [0.1em]
        \includegraphics[width=\linewidth]{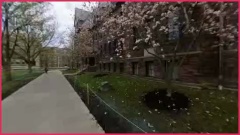}
        }
    \end{minipage}
    \hfill
    \begin{minipage}[t]{0.3786\textwidth}
        \centering
        \small \modelname~(ours) \\ [0.05em]
        \IfDefinedSwitch{\embedPanoVideo}{
        \animategraphics[autoplay,loop,width=\linewidth, trim={0 0 0 0}, clip]{5}{figures-ICCV/comparison-pano/27/ours/}{000}{024} \\ [0.1em]
        \animategraphics[autoplay,loop,width=\linewidth, trim={0 0 0 0}, clip]{5}{figures-ICCV/comparison-pano/44/ours/}{000}{024} 
        }
        {\includegraphics[width=\linewidth]{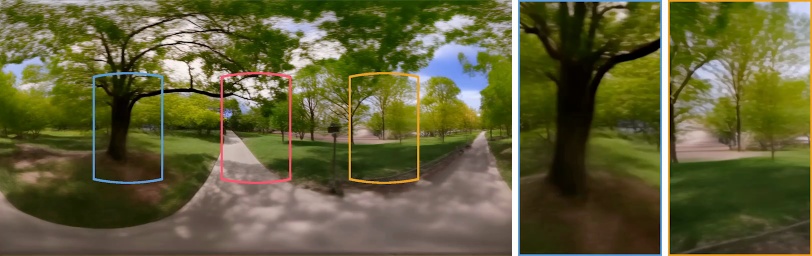} \\ [0.1em]
        \includegraphics[width=\linewidth]{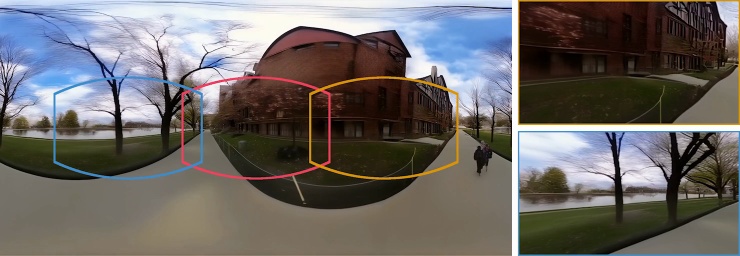}
        }
    \end{minipage}
    \hfill
    \begin{minipage}[t]{0.3786\textwidth}
        \centering
        \small PanoDiffusion~\cite{wu2023panodiffusion} \\ [0.1em]
        \IfDefinedSwitch{\embedPanoVideo}{
        \animategraphics[autoplay,loop,width=\linewidth, trim={0 0 0 0}, clip]{5}{figures-ICCV/comparison-pano/27/pano/}{000}{024} \\ [0.1em]
        \animategraphics[autoplay,loop,width=\linewidth, trim={0 0 0 0}, clip]{5}{figures-ICCV/comparison-pano/44/pano/}{000}{024} 
        }
        {\includegraphics[width=\linewidth]{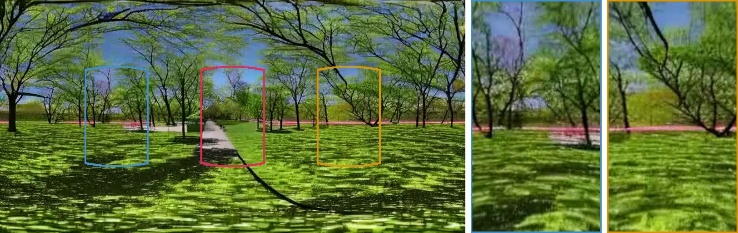} \\ [0.1em]
        \includegraphics[width=\linewidth]{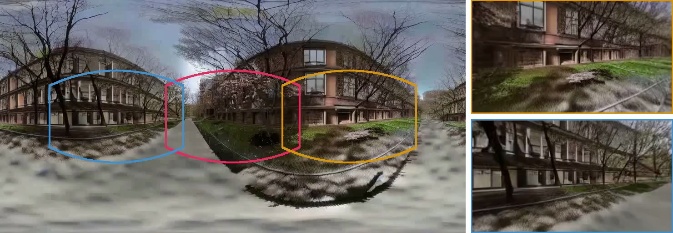}
        }
    \end{minipage}
    \vspace{-6mm}
    \caption{\textbf{Qualitative comparison with 360$^\circ$ image generation method PanoDiffusion (videos embedded).} The input region is highlighted in \textcolor{MyRed}{red}, with \textcolor{MyOrange}{orange} and \textcolor{MyBlue}{blue} regions indicate extracted perspective views. Although PanoDiffusion can generate plausible 360$^\circ$ images from perspective inputs, the generated frames are temporally inconsistent.}
    \label{fig:comparison-panodiffusion}
\end{figure*}

\subsection{Model Training}
\label{sec:model-training}

We train our denoising network $f_\theta$ with a score matching objective:
\begin{align}
\min_{\theta} \, \mathbb{E}_{(\mathbf{x}_{\text{equi}},\mathbf{y}_{\text{equi}})} &_{\sim p_{\text{data}}(\mathcal{E}(X_{\text{equi}}),\mathcal{E}(Y_{\text{equi}})), t, \epsilon \sim \mathcal{N}(0,1)} \notag \\ \
& \lambda(h) ||\epsilon - f_\theta(\mathbf{y}_{\text{equi},t}; t, \mathbf{x}_{\text{equi}}, \mathcal{F}(X_{\text{pers}}))||_2^2.
\end{align}
Here, \( \lambda(h) = (\frac{1}{2} - |\frac{1}{2} - h|)^2 + \delta \) is a re-weighting function that scales the loss of each pixel based on its height $h\in[0,1]$ on the equirectangular map. 
Intuitively, it gives greater importance to regions near the equator (\emph{i.e.}, $h$ closer to $\frac{1}{2}$), as regions near the poles (\emph{i.e.}, $h=0$ or $1$) are disproportionally enlarged in the equirectangular format. $\delta$ is a small offset to ensure that all regions contribute to the loss.

We optimize our model using the EDM~\cite{karras2022elucidating} diffusion framework, parameterizing the denoiser $f_\theta$ as:
\begin{equation}
    f_\theta(\mathbf{y}; \sigma) = c_{\text{skip}}(\sigma) \, \mathbf{y} + c_{\text{out}}(\sigma) \, F_\theta(c_{\text{in}}(\sigma) \, \mathbf{y}; c_{\text{noise}}(\sigma)),
\end{equation}
where $F_\theta$ is the model to be trained, $\sigma=\sigma(t)$ indicates the noise schedule, and $c_{\text{in}}, c_{\text{out}}, c_{\text{skip}}, c_{\text{out}}$ are scaling functions. During training, the noise schedule $\sigma$ is sampled from a log-Gaussian distribution. 
We refer readers to~\cite{karras2022elucidating} for more details on the EDM framework.

\myparagraph{Camera Movement Simulation.}
Our model aims to generate 360$^\circ$ videos from arbitrary perspective videos. 
However, naively sampling perspective views from 360$^\circ$ videos to train diffusion models would be ineffective due to the complex patterns of camera motion in real-world footage. We thus design a sampling strategy that allows us to approximate real-world camera motion and extract realistic training pairs of perspective and 360$^\circ$ videos.

Inspired by~\cite{grundmann2011auto, winter2009biomechanics}, we introduce linear drift, oscillatory, and noise terms to mimic natural human motion~\cite{winter2009biomechanics}.
Formally, camera movement is simulated as follows:
\begin{align}
    \phi_{\text{roll}}(k) &= \mathcal{N}(0, \eta_r) + a_r \sin(\omega k + \tau_r) \notag, \\
    \phi_{\text{pitch}}(k) &= \mathcal{N}(0, \eta_p) + a_p \sin(\omega k+ \tau_p) +  d_p k, \label{eq:camera-motion-simulation} \\
    \phi_{\text{yaw}}(k) &= \mathcal{N}(0, \eta_y) + a_y \sin(\omega k+ \tau_y) +  d_y k + \phi_0, \notag
\end{align}
where $k$ is the frame index, \( \omega \) is the oscillatory frequency, \( \tau_r, \tau_p, \tau_y \) the initial phases, \( a_r, a_p, a_y \) the oscillatory amplitudes, \( \eta_r, \eta_p, \eta_y \) the noise strengths, \( d_p, d_y \) the drift rates, and $\phi_0$ a random offset. The horizontal and vertical field of view are randomly chosen between $[30^\circ, 120^\circ]$. Additionally, since horizontal rotation preserves the $360^\circ$ property, we augment the data with random circular shifts.

\begin{table*}[t!]
\centering
\footnotesize
\scalebox{0.99}{
    \begin{tabular}{lccccccccc}
    \toprule \vspace{0.2mm}
     \multirow{2}{*}{Method} & \multicolumn{3}{c}{$\text{FoV}=60^\circ$} & \multicolumn{3}{c}{$\text{FoV}=90^\circ$}  & \multicolumn{3}{c}{$\text{FoV}=120^\circ$} \vspace{-0.5mm} \\ 
     \cmidrule(lr){2-4}\cmidrule(lr){5-7}\cmidrule(lr){8-10}
     & Imaging$\uparrow$ & Aesthetic$\uparrow$ & Motion$\uparrow$ & Imaging$\uparrow$ & Aesthetic$\uparrow$ & Motion$\uparrow$ & Imaging$\uparrow$ & Aesthetic$\uparrow$ & Motion$\uparrow$ \\ \midrule
     Be-Your-Outpainter~\cite{be-your-outpainter} & 0.4014 & 0.3461 & 0.9683 & 0.4469 & 0.4161 & 0.9649 & 0.4175 & 0.3951 & 0.9628 \\
     Follow-Your-Canvas~\cite{chen2024follow} & 0.4268 & \textbf{0.4750} & 0.9704 & 0.4267 & 0.4685 & 0.9679 & 0.4130 & 0.4513 & 0.9660 \\ 
     \modelname~(ours) & \textbf{0.4760} & 0.4722 & \textbf{0.9816} & \textbf{0.4773} & \textbf{0.4785} & \textbf{0.9796} & \textbf{0.4895} & \textbf{0.4796} & \textbf{0.9777} \\
    \bottomrule
    \end{tabular}
    }
    \vspace{-2mm}
    \caption{\textbf{Quantitative comparison with video outpainting methods}. \emph{Imaging}, \emph{Aesthetic}, and \emph{Motion} stands for the Imaging Quality, Aesthetic Quality, and Motion Smoothness metrics from VBench~\cite{vbench}.}
    \label{tab:qualitative-video-outpainting}
\end{table*}

\begin{figure*}[t!]
    \vspace{-3mm}
    \includegraphics[width=\linewidth]{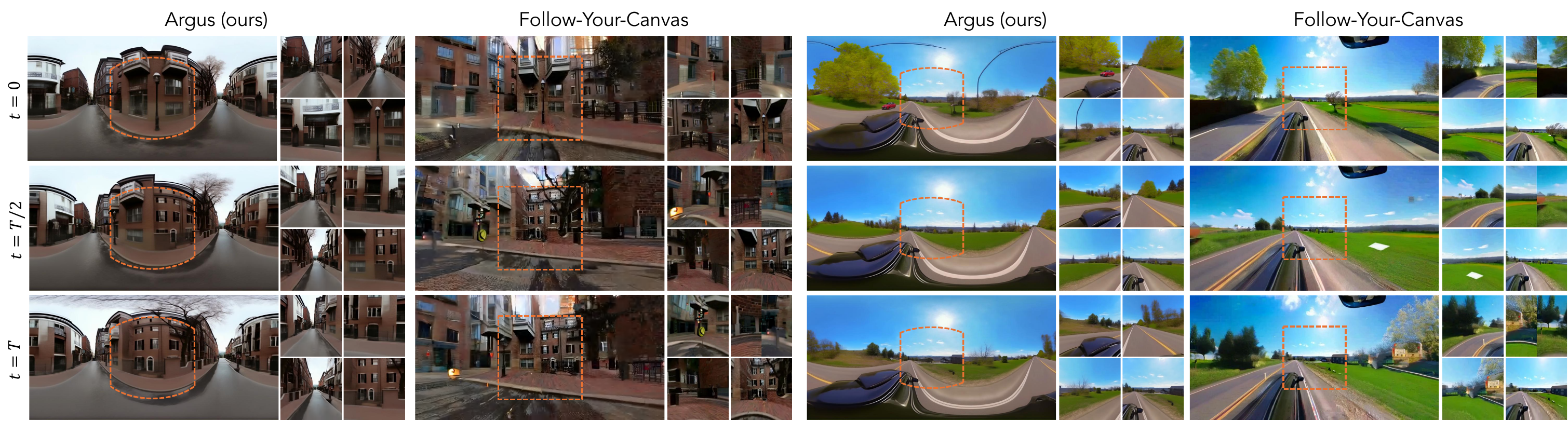}
    \vspace{-6mm}
    \caption{\textbf{Qualitative comparison with state-of-the-art video outpainting method.} 
    The input region is highlighted in \textcolor{MyOrange}{orange}. 
    For each generated 360$^\circ$ frame, four unwrapped perspective views are shown on the right. Video outpainting method struggles with satisfying 360$^\circ$ panoramic property and the generation quality declines as it extends further from the input viewpoint.}
    \vspace{-2mm}
    \label{fig:benchmark-video-outpainting}
\end{figure*}

\subsection{Model Inference}
\label{sec:model-inference}

The above framework is sufficient for training our model on paired 360$^\circ$ and perspective videos. However, generating outputs, especially for in-the-wild videos, presents several challenges. 
First, as discussed in Section~\ref{sec:diffusion}, projecting a perspective video into an equirectangular format typically requires knowledge of the camera’s field of view and poses, yet in practice, the relative camera angles between frames are often unknown.
Another challenge is the presence of boundary artifacts in equirectangular images: while the left and right edges are distant in image space, they are spatially adjacent in the scene. As a result, the model struggles to condition the right edge based on the left and vice versa, causing abrupt changes at the boundary.

\myparagraph{View-Based Frame Alignment.} 
To project the perspectives videos into equirectangular format, one straightforward solution is to always map perspectives frames to the center of equirectangular maps, as shown in Figure~\ref{fig:projection}~(bottom row). While this approach sidesteps the need for camera pose estimation, it forces the diffusion model to implicitly learn the camera motion and handle complex distortions. For example, the model must detect when the camera is panning upward, as in Figure~\ref{fig:projection}~(bottom row), and predict surrounding content according to varying patterns of spherical distortion. Furthermore, the sky may appear in different locations within the 360$^\circ$ scene, further complicating the task. To address this challenge, we first estimate the relative camera poses of the input video using SLAM framework~\cite{li2024megasam}. We then compute the Euler angles relative to the first frame and project them onto the equirectangular map. 
As shown in Figure~\ref{fig:projection}~(middle row), this coordinate alignment ensures that each part of the equirectangular map corresponds to roughly the same scene region across frames, significantly improving consistency. For example, the sky appears consistently at the top, while the road remains at the bottom.

\begin{figure*}[t!]
    \includegraphics[width=\linewidth]{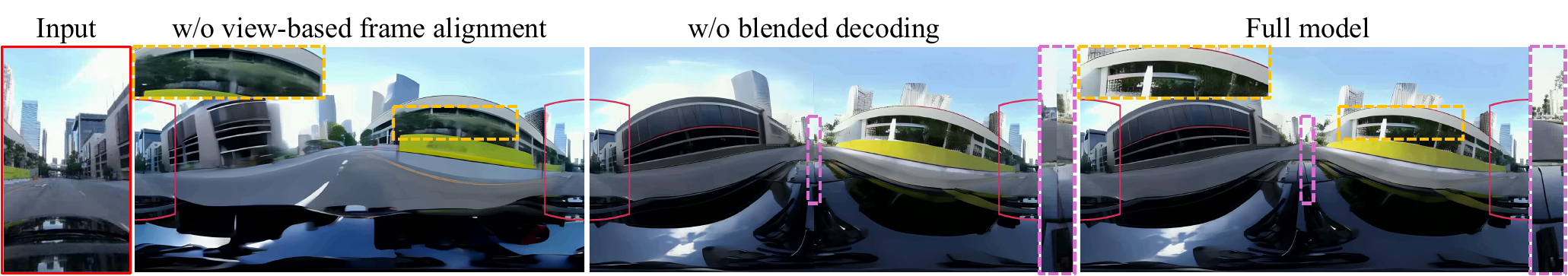}
    \vspace{-6mm}
    \caption{\textbf{Qualitative ablation studies.} The input region is marked in \textcolor{MyRed}{red}. The 360$^\circ$ images are rotated 180$^\circ$ to illustrate the panoramic consistency. Compared to our full model, the variant without view-based frame alignment appears blurrier (\textcolor{MyOrange}{orange} box), while the variant without blended decoding shows artifacts in the center (\textcolor{MyPink}{pink} box). Boxes are enlarged for ease of visualization.}
    \label{fig:ablation}
\end{figure*}

\begin{figure*}[t!]
    \centering
    \vspace{-1mm}
    \scalebox{1}{
        \begin{minipage}[t]{0.495\textwidth}
            \centering
            \IfDefinedSwitch{\embedInthewildVideo}{\animategraphics[autoplay,loop,width=\linewidth, trim={0 0 0 0}, clip]{5}{figures-ICCV/in-the-wild/3/}{000}{044}
            }
            {\includegraphics[width=\linewidth]{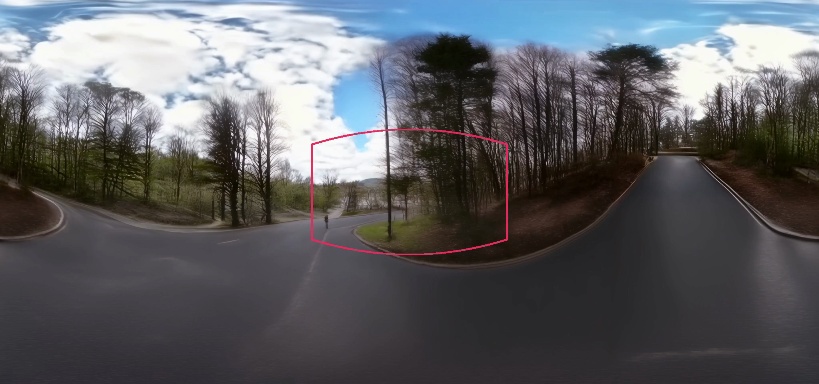}
            }
        \end{minipage}
        \hfill
        \begin{minipage}[t]{0.495\textwidth}
            \centering
            \IfDefinedSwitch{\embedInthewildVideo}{\animategraphics[autoplay,loop,width=\linewidth, trim={0 0 0 0}, clip]{5}{figures-ICCV/in-the-wild/45/}{000}{044}
            }
            {\includegraphics[width=\linewidth]{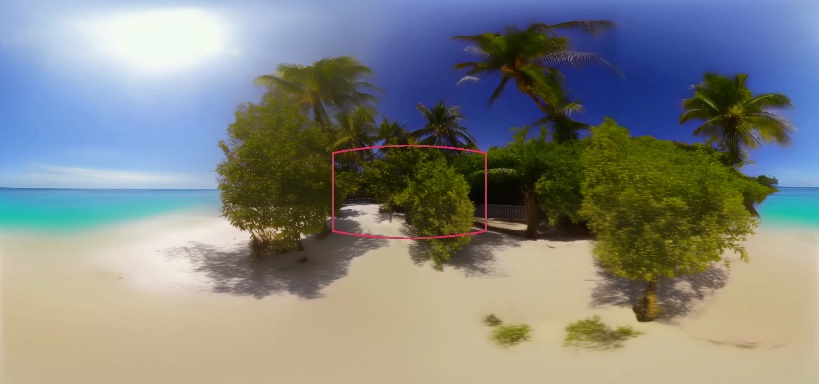}
            }
        \end{minipage}
    }
    \vspace{-6mm}
    \caption{\textbf{Long-term 360$^\circ$ video generation in the wild.} The input video region is marked in \textcolor{MyRed}{red}. Our generated results maintain semantic consistency across two rounds of generation. View the video results on our~\projectpagehref.
    \vspace{-2mm}
    }
    \label{fig:in-the-wild}
\end{figure*}

\myparagraph{Blended Decoding.} When generating 360$^\circ$ video frames, inconsistencies often emerge at the boundary where the left and right edges of the equirectangular image meet. To address this, we introduce blended decoding (Figure~\ref{fig:blend-decoding}).

Previous techniques such as two-end alignment sampling~\cite{wu2023panodiffusion} and circular padding~\cite{360dvd} operate in the latent space, which cannot guarantee smooth boundary transitions after decoding, as the VAE is trained on standard perspective images or videos only. We propose blending in the pixel space instead. Specifically, we decode both the original latent and a 180$^\circ$-rotated version, creating two outputs with identical content but differently positioned artifacts. We then compute a distance-based weighted average, assigning greater weight to pixels farther from the boundary:
\begin{equation}
Y_{k,i,j} = h_W(i)Y_{k,i,j} + (1-h_W(i))Y'_{k,i,j},
\end{equation}
\begin{equation}
h_W(x) = 1 - 2\left|\frac{x}{W} - \frac{1}{2}\right|.
\end{equation}
Here, $i$ and $j$ refer to the pixel coordinates. $Y_k$ and $Y'_k$ denote the equirectangular frames generated at 0$^\circ$ and 180$^\circ$ offsets for frame $k$. $W$ represents the image width. This approach allows us to blend the two videos, effectively mitigating boundary artifacts. See Figure~\ref{fig:blend-decoding} for qualitative examples.

\myparagraph{Long Video Generation.} {The method described above is limited to generating 360$^{\circ}$ panoramas from input perspective videos of exactly $T$ frames. To accommodate longer input sequence, we extend our approach through context-aware training. Concretely, the model learns to predict the subsequent $T-S$ frames conditioned on $S$ initial frames, which are fully observable in the conditioning equirectangular video. During training, we alternate between standard inputs (all $T$ conditioning frames masked) and context-aware inputs (first $S$ frames visible, remaining $T-S$ frames masked). For inference on extended sequences, we implement an iterative sampling process in which recent predictions serve as a context for subsequent iterations, allowing the generation of longer-length panoramic videos.}

%% file: sec/5_experiments.tex
\section{Experiments}
\label{sec:exp}

In this section, we first present a quantitative evaluation of~\modelname, followed by qualitative examples of 360$^\circ$ generation from in-the-wild videos. Finally, we present a diverse set of downstream tasks that \modelname~can be applied to off-the-shelf.

\subsection{Experimental Setup}
Our model is initialized from the Stable Video Diffusion-I2V-XL model~\cite{blattmann2023stable}. We train it in two phases: first at $384\times768$ resolution for 100K iterations, then finetuning on a high-quality subset at $512\times1024$ resolution for additional 20K iterations, both with batch size 16. The finetuning phase adopts context-aware training and employs a noisier distribution to enhance training effectiveness at higher resolutions~\cite{chen2023importance}. We set sequence length $T=25$ and context length $S=5$.  
We briefly describe our data, metrics, and baselines below, with complete details available in the supp. material.

\begin{figure*}[t]
    \vspace{-2mm}
    \centering
    \begin{minipage}[t]{0.33\textwidth}
        \centering
        \small Input Video \\ [0.1em]
        \IfDefinedSwitch{\embedStabilizationVideo}{
        \animategraphics[autoplay,loop,width=\linewidth, trim={0 0 0 0}, clip]{5}{figures-ICCV/stabilization/20/input/}{000}{024}
        }{
        \includegraphics[width=\linewidth]{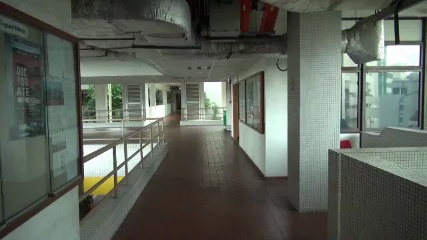}
        }
    \end{minipage}
    \hfill
    \begin{minipage}[t]{0.33\textwidth}
        \centering
        \small Stabilization (\modelname) \\ [0.1em]
        \IfDefinedSwitch{\embedStabilizationVideo}{
        \animategraphics[autoplay,loop,width=\linewidth, trim={0 0 0 0}, clip]{5}{figures-ICCV/stabilization/20/ours/}{000}{024}
        }{
        \includegraphics[width=\linewidth]{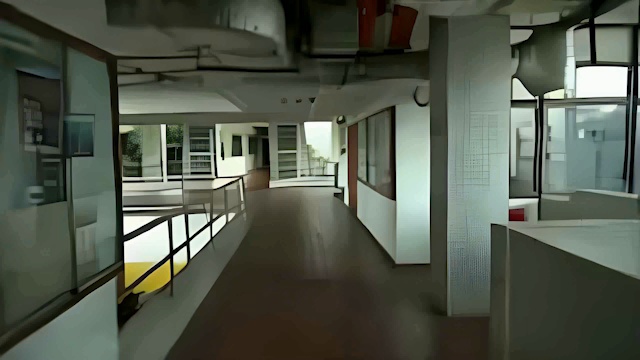}
        }
    \end{minipage}
    \hfill
    \begin{minipage}[t]{0.33\textwidth}
        \centering
        \small Stabilization (reference) \\ [0.1em]
        \IfDefinedSwitch{\embedStabilizationVideo}{
        \animategraphics[autoplay,loop,width=\linewidth, trim={0 0 0 0}, clip]{5}{figures-ICCV/stabilization/20/reference/}{000}{024}
        }{
        \includegraphics[width=\linewidth]{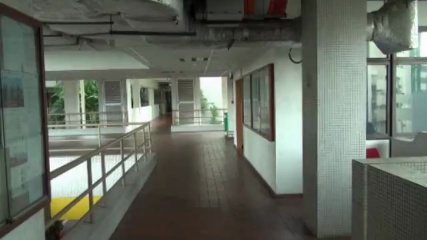}
        }
    \end{minipage}
    \vspace{-6mm}
    \caption{\textbf{Video stabilization results (videos embedded).} Columns from left to right: input frames, result from \modelname, and reference result from~\cite{liu2013bundled}. Unlike cropping-based approaches, \modelname~maintains the full field of view due to its panoramic generation capability.}
    \vspace{-1mm}
    \label{fig:video-stabilization}
\end{figure*}

\begin{table}[t!]
\centering
\scriptsize
    \scalebox{0.80}{
        \begin{tabular}{lcccccc}
        \toprule \vspace{0.05mm}
         Variant & PSNR$\uparrow$ & LPIPS$\downarrow$ & FVD$\downarrow$ & Imaging$\uparrow$ & Aesthetic$\uparrow$ & Motion$\uparrow$ \\
        \midrule
        w/o frame alignment & 20.42 & 0.3194 & 1349.6 & 0.3816 & 0.4604 & 0.9783 \\
        w/o blended decoding &  \textbf{22.09} & 0.2675 & \textbf{1226.3} & 0.4574 & 0.4705 & 0.9795 \\
        Full model & 21.83 & \textbf{0.2409} & 1228.6 & \textbf{0.4939} & \textbf{0.4828} & \textbf{0.9802} \\ \midrule
        VAE Reconstruction & 24.54 & 0.1663 & 121.8 & 0.5272 & 0.4929 & 0.9793 \\
        \bottomrule
        \end{tabular}
    }
    \vspace{-1mm}
    \caption{\textbf{Ablation studies.} Our view-based frame alignment technique significantly improves overall performance, while blended decoding notably enhances boundary consistency despite its minimal effect on quantitative scores. Results of direct reconstruction using VAE are listed to represent the performance upper bound.}
    \vspace{-2mm}
\label{tab:ablation}
\end{table}

\begin{figure}[t!]
    \centering
    \vspace{-3mm}
    \begin{minipage}[t]{0.155\textwidth}
        \centering
        \scriptsize Input Video \\ 
        \IfDefinedSwitch{\embedRotationVideo}{\animategraphics[autoplay,loop,width=\linewidth, trim={0 0 0 0}, clip]{5}{figures-ICCV/camera-control/47/input/}{001}{024} \\ [0.15em]
        \animategraphics[autoplay,loop,width=\linewidth, trim={0 0 0 0}, clip]{5}{figures-ICCV/camera-control/43/input/}{001}{024}
        }{\includegraphics[width=\linewidth]{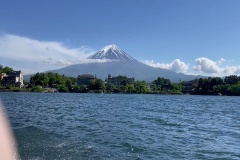} \\ [0.15em]
        \includegraphics[width=\linewidth]{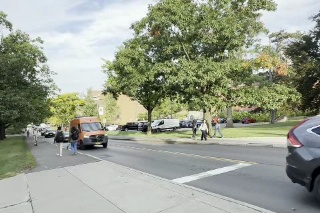}
        }
    \end{minipage}
    \hfill
    \begin{minipage}[t]{0.155\textwidth}
        \centering
        \scriptsize Rotate 30° clockwise \\ [0.2em]
        \IfDefinedSwitch{\embedRotationVideo}{\animategraphics[autoplay,loop,width=\linewidth, trim={0 0 0 0}, clip]{5}{figures-ICCV/camera-control/47/rotate30/}{001}{024} \\ [0.15em]
        \animategraphics[autoplay,loop,width=\linewidth, trim={0 0 0 0}, clip]{5}{figures-ICCV/camera-control/43/rotate30/}{001}{024}
        }{\includegraphics[width=\linewidth]{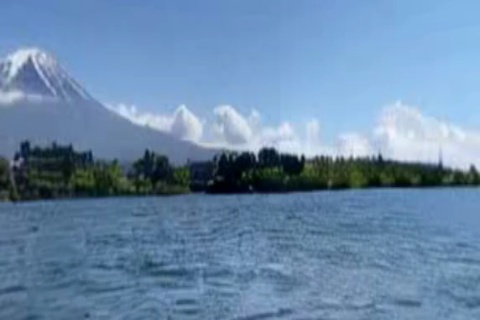} \\ [0.15em]
        \includegraphics[width=\linewidth]{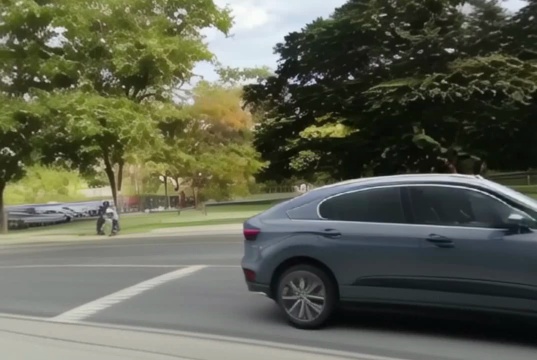}
        }
    \end{minipage}
    \hfill
    \begin{minipage}[t]{0.155\textwidth}
        \centering
        \scriptsize Rotate 45° clockwise \\ [0.2em]
        \IfDefinedSwitch{\embedRotationVideo}{\animategraphics[autoplay,loop,width=\linewidth, trim={0 0 0 0}, clip]{5}{figures-ICCV/camera-control/47/rotate45/}{001}{024} \\ [0.15em]
        \animategraphics[autoplay,loop,width=\linewidth, trim={0 0 0 0}, clip]{5}{figures-ICCV/camera-control/43/rotate45/}{001}{024}
        }{\includegraphics[width=\linewidth]{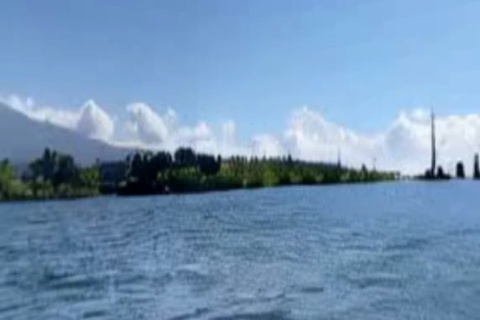} \\ [0.15em]
        \includegraphics[width=\linewidth]{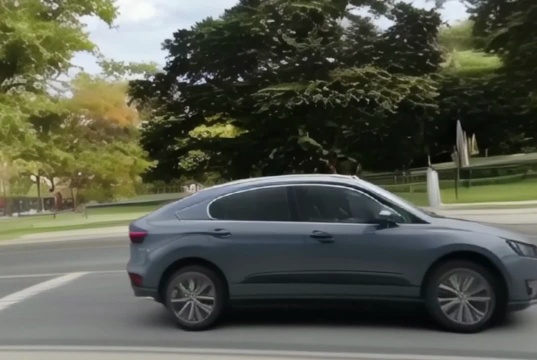}
        }
    \end{minipage}
    \vspace{-6mm}
    \caption{\textbf{Camera control in dynamic scenes (videos embedded).} Our model enables free camera rotation within dynamic scenes to capture elements beyond the initial viewpoint.}
    \vspace{-3mm}
    \label{fig:camera-control}
\end{figure}

\myparagraph{Data.} We evaluate our approach using a dataset of 101 360$^\circ$ videos, captured either with Insta360 cameras or from a hold-out set from YouTube. The 360$^\circ$-perspective video pairs are created using two types of camera trajectories: (i) simulated trajectories generated by our camera motion simulation technique, and (ii) real-world trajectories extracted through calibration. Additionally, we collected 15 videos featuring linear structures, such as lanes and sidewalks, to evaluate geometric consistency in extrapolated views.

\myparagraph{Metrics.} {We evaluate our results based on three key criteria: image quality, temporal coherency, and geometric consistency. For image quality, we use PSNR, LPIPS~\cite{lpips}, Imaging Quality, and Aesthetic Quality metrics from VBench~\cite{vbench}. For temporal coherency, we employ FVD~\cite{unterthiner2019fvd} and Motion Smoothness~\cite{vbench}. For geometric consistency, we introduce a \emph{line consistency} metric to evaluate whether straight lines remain straight within extrapolated views. This metric is particularly important for assessing whether our model preserves fundamental geometric properties when generating novel views. To quantitatively measure this consistency, we follow~\cite{qian2023understanding} and use EA-score~\cite{zhao2021deephough} to evaluate the angular and Euclidean distances between line pairs.}

\myparagraph{Baselines.} Since no existing method is explicitly designed for the video-to-360$^\circ$ task, we adapt PanoDiffusion~\cite{wu2023panodiffusion}, a 360$^\circ$ image generation method, as our first baseline. Specifically, we re-trained their model on 360$^\circ$ video frames from our dataset without the depth branch. 
To improve consistency across frames, we  applied identical initial noise across all frames during the sampling process~\cite{song2020denoising}. 
We also compare~\modelname~with video outpainting methods~\cite{chen2024follow, be-your-outpainter}. Since these baselines support only rectangular input, we center square videos on the canvas and expand the vertical and horizontal field of view (FoV) to 180$^\circ$ and 360$^\circ$, respectively. For evaluation, we extracted three perspective videos from each 360$^\circ$ test video, with FoVs of 60, 90, and 120 degrees.

\subsection{Results and Analyses}

\myparagraph{Quantitative and Qualitative Results.}
We evaluate our model and baselines on our curated 360$^\circ$-perspective video pairs. We use GT camera trajectories for all methods to isolate the impact of imperfect camera poses. 
As shown in Table~\ref{tab:comparison-panodiffusion} and Figure~\ref{fig:comparison-panodiffusion}, \modelname~significantly outperforms the adapted PanoDiffusion. 
While the adapted PanoDiffusion generates plausible individual 360$^\circ$ frames, it struggles with temporal consistency.  
\modelname~, in contrast, produces temporally smooth results, and is able to understand the geometric layout in the input and correctly extrapolate beyond. 
Comparing with video outpainting baselines, our method also achieves better visual quality and temporal coherency (see Table~\ref{tab:qualitative-video-outpainting} and Figure~\ref{fig:benchmark-video-outpainting}).
Video outpainting methods notably fail to preserve 360$^\circ$ panoramic properties, with generation quality deteriorating as the distance from the original viewpoint increases. In contrast, our model produces realistic panoramic videos throughout the entire field of view.

\myparagraph{Ablation Studies.} 
To verify the effectiveness of view-based frame alignment, we train a model in which perspective videos are always centered within the equirectangular map. During evaluation, we adjust the GT 360$^\circ$ videos accordingly. As shown in Table~\ref{tab:ablation} and Figure~\ref{fig:ablation}, the absence of viewpoint alignment leads to degraded performance. This supports our hypothesis in Section \ref{sec:model-inference} that without viewpoint alignment, the diffusion model must implicitly learn camera motion and manage complex distortions, making the task significantly more challenging.
Table~\ref{tab:ablation} also showcases the importance of blended decoding. 
For reference, we include results from direct reconstruction using the VAE, which represents the performance upper bound.

\begin{figure}[t]
    \centering
    \vspace{-3mm}
    \includegraphics[width=\linewidth]{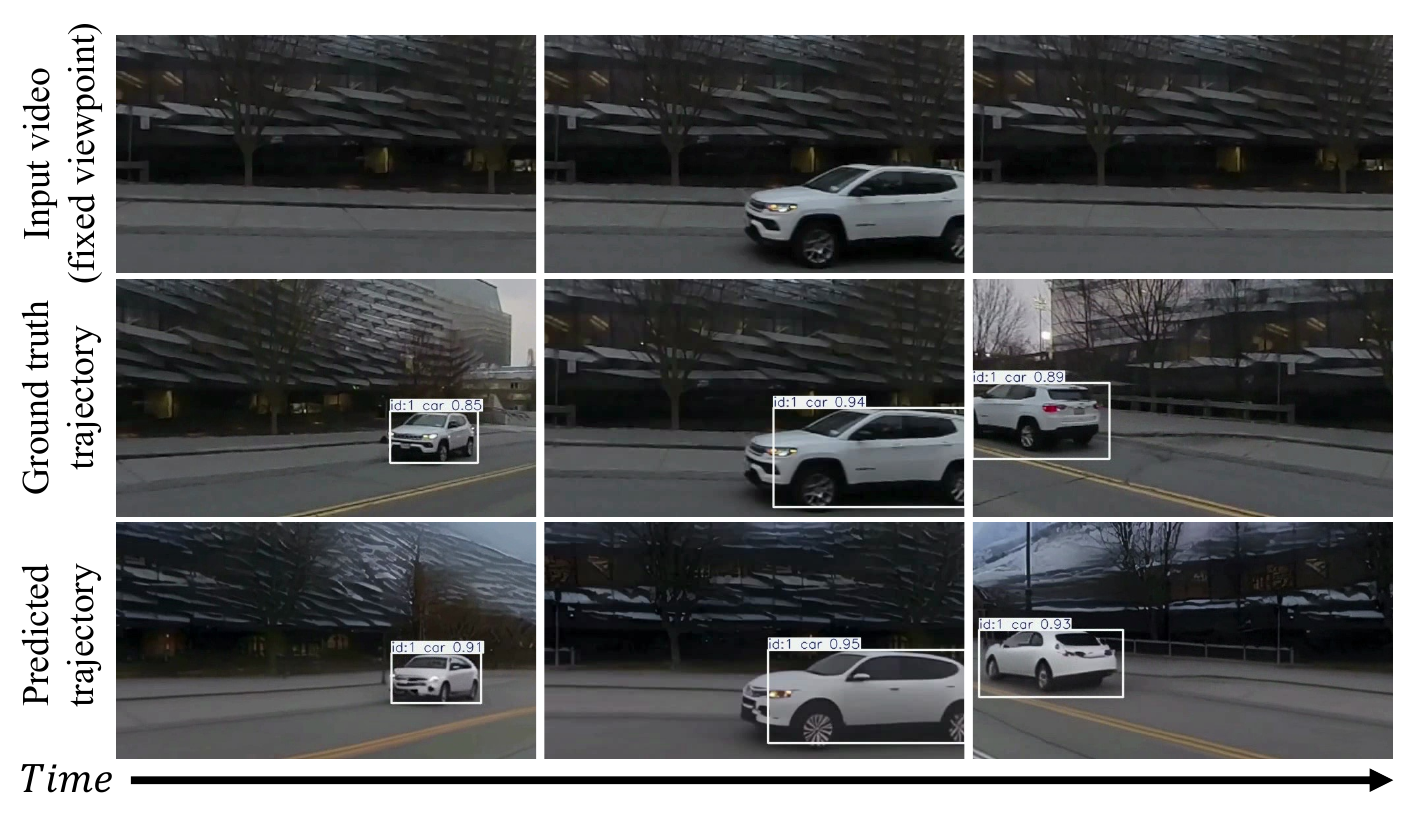}
    \vspace{-7mm}
    \caption{\textbf{Interpreting scene dynamics.} We capture a car driving scene with 360$^\circ$ camera and provide our model with a 60$^\circ$ FoV input of fixed viewing direction (top). The car's ground truth trajectory (middle) and Argus's predicted trajectory (bottom) shows strong alignment, demonstrating its ability to accurately predict object dynamics beyond the visible field of view.}
    \label{fig:dynamics}
    \vspace{-5mm}
\end{figure}

\begin{figure*}[!t]
    \centering
    \includegraphics[width=\linewidth]{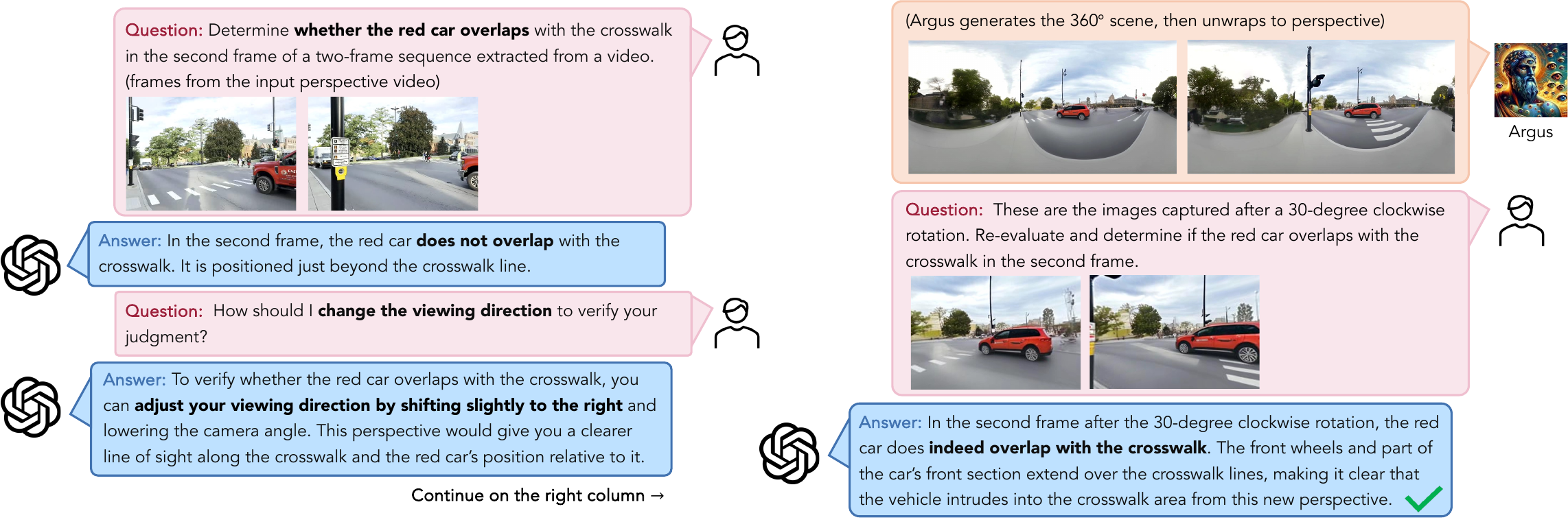}
    \vspace{-5mm}
    \caption{\textbf{Interactive visual question answering.} The first image sequence shows a red vehicle approaching a crosswalk, where the vision-language model (GPT-4o) fails to answer the question correctly because it lacks full scene comprehension. With~\modelname, we can freely rotate the camera, enabling better spatial understanding and accurately revealing the vehicle’s overlap with the crosswalk.}
    \label{fig:application-VQA}
\end{figure*}

\myparagraph{360$^\circ$ Video Generation In the Wild.}
Besides the curated 360$^\circ$-perspective video pairs, we test our model on in-the-wild perspective videos featuring a diverse range of camera motions and environments.
We calibrate camera poses and employ iterative sampling for extended video generation. 
Our model is able to handle fixed orientation (Figure~\ref{fig:in-the-wild}, left), mild motion (Figure~\ref{fig:in-the-wild}, right), rapid motion (Figure~\ref{fig:teaser}), panning and vertical movement (\projectpagehref), and even synthetic inputs from a text-to-video model (\projectpagehref).

\myparagraph{Interpreting Scene Dynamics.} 
As we have alluded to in Figure \ref{fig:teaser}, our model can understand the dynamics encoded in the input video (\emph{e.g.}, the motion of the car) and extrapolate beyond. To better evaluate whether the generated dynamics are reasonable, we first capture a 360$^\circ$ video of a car driving by. We then crop a 60° horizontal FoV and input it into \modelname. Finally, we apply tracking to both the generated 360$^\circ$ video and the original footage. As shown in Figure~\ref{fig:dynamics}, the predicted trajectory closely aligns with the car's ground-truth motion. See our \projectpagehref~for more details.

\myparagraph{Scene Generation Plausibility.} Beyond line consistency, we evaluate the geometric plausibility of our generated 360° videos through 3D reconstruction. We unwrap panning perspective videos with yaw angles ranging from 45° to -45°, then calibrate using MegaSaM~\cite{li2024megasam}. Testing on 48 smartphone videos, we compare predicted rotation angles with unwrapping angles and observe minimal average differences of $(\Delta\text{roll},\Delta\text{pitch},\Delta\text{yaw})=(0.22^\circ,0.30^\circ,0.34^\circ)$, confirming our generated content achieves high geometric realism.

\begin{figure}[t]
    \vspace{-2.5mm}
    \centering
    \includegraphics[width=\linewidth]{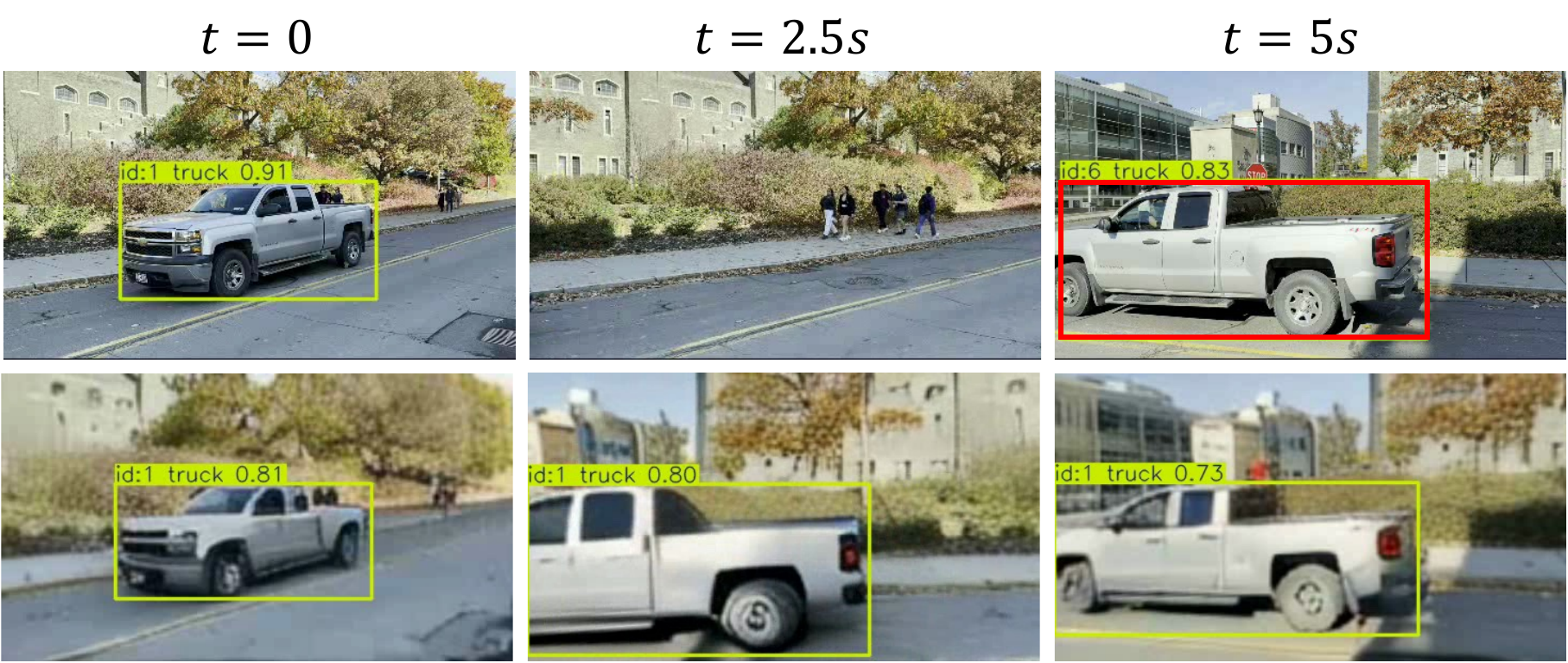}
    \vspace{-6mm}
    \caption{\textbf{Consistent object tracking.} Object detection results comparing input video (top) versus our unwrapped panorama (bottom). While the truck is identified as a separate entity when exiting and re-entering the input frame, it remains continuously visible in our generated panorama, resulting in consistent tracking.}
    \label{fig:tracking}
    \vspace{-5mm}
\end{figure}

\subsection{Applications}

This section showcases \modelname’s potential applications, including video stabilization, camera viewpoint control, dynamic environmental mapping, and interactive VQA.

\myparagraph{Video Stabilization.} \modelname~shows promising application to video stabilization without modifications. Traditional video stabilization techniques require cropping, resulting in a reduced field of view and visual information loss. In contrast, \modelname~enables video stabilization while maintaining a consistent field of view, as the generated panorama preserves scene information across frames.
To achieve higher-resolution outputs, we crop regions with a smaller field of view from 360$^\circ$ videos and finetune on them. 
We test our approach using the video stabilization dataset from~\cite{liu2013bundled}. As shown in Figure~\ref{fig:video-stabilization}, our method produces visually pleasing stabilization results while preserving a larger field of view than the reference results, effectively overcoming the limitations of cropping.

\myparagraph{Camera Viewpoint Control.} \modelname~enables viewpoint control in dynamic environments by unwrapping the generated 360° scene into perspective views. This capability allows exploration beyond the initial field of view (Figure~\ref{fig:camera-control}) and facilitates tracking of fast-moving objects (Figure~\ref{fig:tracking}), enhancing immersion and supporting scene understanding tasks.

\myparagraph{Dynamic Environmental Mapping.} \modelname~enables realistic object relighting using the generated 360° panorama videos as dynamic environment maps. Figure~\ref{fig:teaser} showcases metallic spheres rendered with these videos, exhibiting accurate reflections and lighting that validate practical applications.

\myparagraph{Interactive VQA.} Finally, we explore how the generated panorama video can help visual question answering in dynamic environments. Although generated videos might not provide a solid ground of facts, we show that by enabling free rotation of the camera, \modelname~allows for comprehensive spatial understanding by seeing the scene from multiple perspectives, based on the signals fully or partially available within the input perspectives. This flexibility supports interactive visual question answering, such as verifying if a vehicle overlaps with a crosswalk (Figure~\ref{fig:application-VQA}). This capability overcomes the limitation of fixed-viewpoint videos and enhances scene comprehension and opens new possibilities for video analysis applications.

%% file: sec/6_conclusion.tex
\section{Discussion}
\myparagraph{Limitations.}  
Due to computational resource constraints, our current output resolution ($512\times 1024$) is lower than that of typical 4K real-world panoramas. The resolution further decreases when unwrapping back to perspective views. Additionally, while our model substantially improves upon the base SVD model in terms of object dynamics and temporal consistency (see supp. material for comparisons), it still exhibits shape inconsistencies and physics artifacts, similar to SVD and other SoTA video models such as COSMOS.

\myparagraph{Conclusion.} 
We present \modelname, a video-to-360$^\circ$ generation model that creates full 360$^\circ$ panoramas from single-view perspective videos. \modelname~is trained on a relatively untapped data source, 360$^\circ$ videos. To enhance 360$^\circ$ video generation, we incorporate techniques such as camera movement simulation, blended decoding, and view-based frame alignment. \modelname~demonstrates strong performance across varied video sources, effectively capturing dynamic scenes with seamless spatial continuity. Our model offers promising potential for a broad range of downstream applications, marking a step forward in panoramic video generation.

\myparagraph{Acknowledgment.} The research is partially supported by a gift from Ai2, NVIDIA Academic Grant, and DARPA TIAMAT program No. HR00112490422. Its contents are solely the responsibility of the authors and do not necessarily represent the official views of DARPA.

%% file: sec/7_supplementary_arxiv.tex
\clearpage
\setcounter{page}{1}
\setcounter{section}{0}
\maketitlesupplementary

\section{Supplementary Material Overview}
In this supplementary material, we provide additional dataset and implementation details. Accompanying this supplementary file is our~\projectpagehref. 

\section{Dataset Collection and Statistics}

While 360° videos have been utilized on a small scale for various vision applications~\cite{sun360, cai2021extreme,bezalel2024extreme}, their potential remains largely unexplored at greater magnitudes. In this section, we introduce a scalable data curation strategy for training a video-to-360$^\circ$ diffusion model. Then we show examples from our dataset and introduce its statistics to provide a rough understanding of our dataset.

\subsection{Data Processing}

We begin with the 360-1M dataset~\cite{wallingford2024imagine}, which includes approximately 1 million 360$^\circ$ videos of varying quality. To establish a quality baseline, we retain only videos with more than 50 likes. Despite this initial filtering, the dataset still contains mislabeled 180$^\circ$ videos, standard perspective videos, static posters, static scenes, and unrealistic animations. To address this, we developed a scalable data processing pipeline:

\begin{enumerate}
    \item \textbf{Format Filtering.} We sample frames from each video and detect horizontal lines in the center or vertical lines at the boundaries to verify the equirectangular format. Horizontal line detection removes up-down formatted 360$^\circ$ videos, while vertical line detection filters out perspective videos and posters.
    \item \textbf{Intra-frame Filtering.} We compute LPIPS between the left and right halves to filter 180$^\circ$ videos and between the top and bottom halves to filter improperly formatted 360$^\circ$ videos.
    \item \textbf{Inter-frame Filtering.}  To ensure scene dynamics, we sample frames at random intervals and calculate the pixel variance. Static videos with minimal inter-frame variation are removed.
\end{enumerate}

After coarse filtering, the videos are split into 10-second clips. We then apply fine-grained filtering using optical flow~\cite{farneback2003two} to detect low-motion clips, TransNetv2~\cite{soucek2024transnet} to identify cuts, and DPText-DETR~\cite{ye2023dptext} to detect texts from unwrapped perspective views. Clips with excessive black pixels or low pixel variance are also excluded, as they indicate low visual complexity.

\begin{figure}[t]
    \centering
    \includegraphics[width=\linewidth]{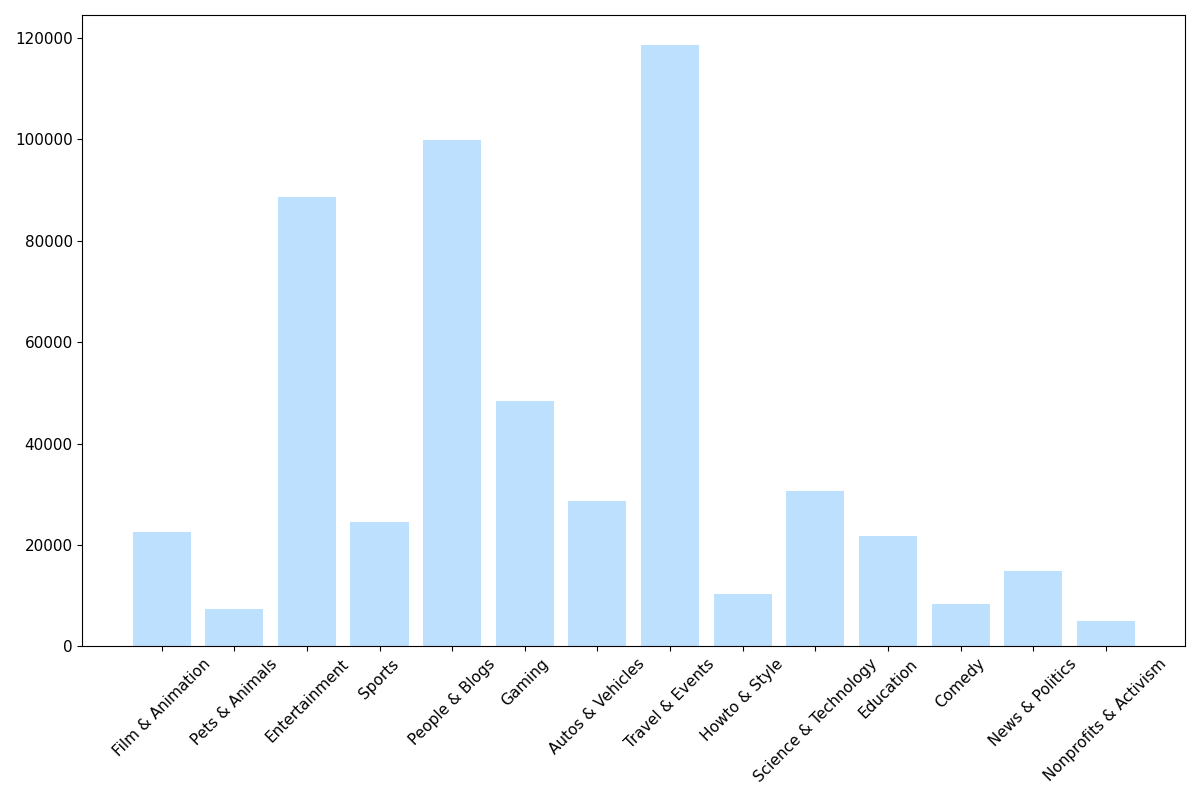}
    \vspace{-7mm}
    \caption{\textbf{Clip category distribution in our dataset.} }
    \label{fig:dataset-distribution}
    \vspace{2mm}
\end{figure}

\subsection{Dataset Statistics}

\begin{figure}[t]
    \centering
    \includegraphics[width=\linewidth]{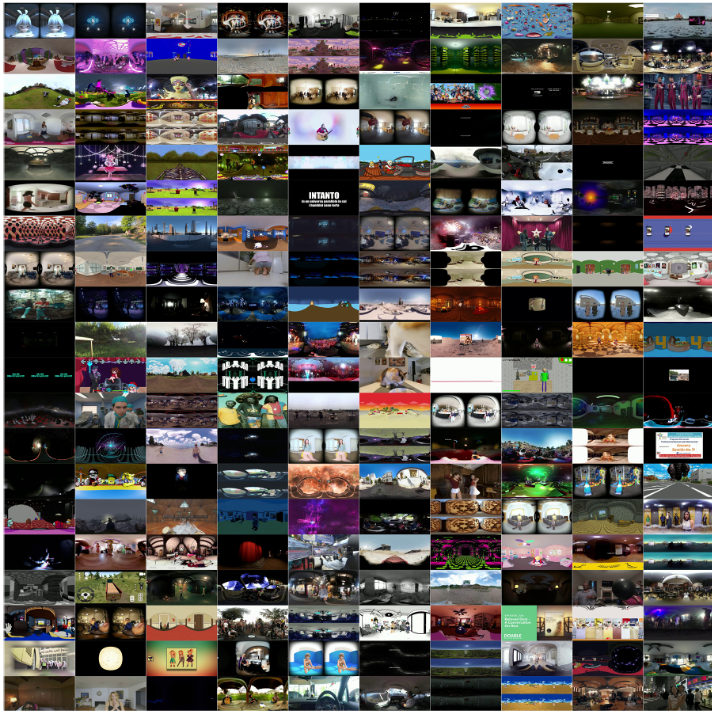}
    \vspace{-4mm}
    \caption{\textbf{Examples of videos discarded during data the data filtering pipeline.} We discard 180$^\circ$ videos, standard perspective videos, static posters, static scenes, and unrealistic animations from the initial noisy dataset.}
    \label{fig:dataset-bad}
\end{figure}

\begin{figure*}[t]
    \centering
    \includegraphics[width=\linewidth]{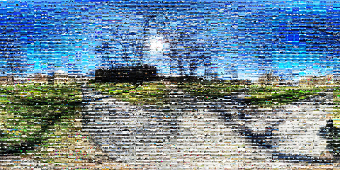}
    \vspace{-5mm}
    \caption{\textbf{Video frames sampled from our dataset.} We arrange the video frames to from a 360$^\circ$ image.}
    \vspace{-3mm}
    \label{fig:dataset-good}
\end{figure*}

The final dataset consists of 283,863 ten-second clips, distributed across 14 subject categories. The most prominent category, ``Travel and Events,'' accounts for 63,935 clips. From this dataset, we also build a high-quality selected after manual inspection of the video frames. This subset was used for high-quality fine-tuning. The distribution of categories in the dataset is shown in Figure~\ref{fig:dataset-distribution}, with examples of filtered and included clips in Figures~\ref{fig:dataset-bad} and~\ref{fig:dataset-good}.

\section{Implementation Details and Analyses}

\subsection{Perspective to Equirectangular Projection}
\label{sec:supp-projection}

We detail the mathematical process of mapping perspective video pixels to equirectangular maps. This includes equations for coordinate normalization, rotation, and spherical mapping.

To map a pixel coordinate $(u, v)$ from an image with a given field of view, roll, pitch, and yaw to an equirectangular map, we first normalize the pixel coordinates to the normalized device coordinates (NDC). Assuming an image resolution of $(W, H)$, the NDC coordinates $(x_{ndc}, y_{ndc})$ are given by
\begin{equation}
x_{ndc} = \frac{2u}{W} - 1, \quad y_{ndc} = \frac{2v}{H} - 1.
\end{equation}
Given horizontal and vertical FOVs $\alpha$ and $\beta$, we compute a 3D direction vector $(X, Y, Z)$ for the pixel in the camera’s coordinate frame as follows:
\begin{equation}
X = x_{ndc} \cdot \tan\left(\frac{\alpha}{2}\right), \quad Y = y_{ndc} \cdot \tan\left(\frac{\beta}{2}\right), \quad Z = -1.
\end{equation}
To reorient this vector from the camera frame to the equirectangular frame, we apply a series of rotations defined by the roll $r$, pitch $p$, and yaw $y$ angles. Each angle defines a rotation matrix: $R_r$ for roll,
\begin{equation}
R_r = \begin{bmatrix}
1 & 0 & 0 \\
0 & \cos(r) & -\sin(r) \\
0 & \sin(r) & \cos(r)
\end{bmatrix},
\end{equation}
$R_p$ for pitch,
\begin{equation}
R_p = \begin{bmatrix}
\cos(p) & 0 & \sin(p) \\
0 & 1 & 0 \\
-\sin(p) & 0 & \cos(p)
\end{bmatrix},
\end{equation}
and $R_y$ for yaw,
\begin{equation}
R_y = \begin{bmatrix}
\cos(y) & -\sin(y) & 0 \\
\sin(y) & \cos(y) & 0 \\
0 & 0 & 1
\end{bmatrix}.
\end{equation}
The rotated vector $(X', Y', Z')$ is obtained by applying these transformations in the order $R_y \cdot R_p \cdot R_r$:
\begin{equation}
\begin{bmatrix}
X' \\
Y' \\
Z'
\end{bmatrix}
= R_y \cdot R_p \cdot R_r \cdot
\begin{bmatrix}
X \\
Y \\
Z
\end{bmatrix}.
\end{equation}
We then convert $(X', Y', Z')$ to spherical coordinates, where $\theta = \arctan2(Y', X')$ and $\phi = \arcsin\left(\frac{Z'}{\sqrt{X'^2 + Y'^2 + Z'^2}}\right)$. Finally, the spherical coordinates are mapped to equirectangular pixel coordinates $(u_{eq}, v_{eq})$ for an equirectangular map of dimensions $(W_{eq}, H_{eq})$ by
\begin{equation}
u_{eq} = \frac{W_{eq}}{2\pi} \cdot (\theta + \pi), \quad v_{eq} = \frac{H_{eq}}{\pi} \cdot \left(\frac{\pi}{2} - \phi\right).
\end{equation}
This yields the pixel location on the equirectangular map corresponding to the input pixel in the original image.

\subsection{Training Details}
Our model is initialized from the Stable Video Diffusion-I2V-XL model~\cite{blattmann2023stable}. We implement a two-phase training strategy: initially at $384\times768$ resolution for 100K iterations, where we sample the noise scheduler parameter $\sigma$ from a log-Gaussian distribution ($\log\sigma\sim\mathcal{N}(P_{\text{mean}}, P_{\text{std}}^2)$) and progressively increase the noise schedule from $(P_{\text{mean}}, P_{\text{std}}^2) = (-1, 1)$ to $(0, 1)$. In the second phase, we finetune the model at higher $512\times 1024$ resolution on a high-quality subset for 20K iterations, employing context-aware training with a stronger noise schedule of $(P_{\text{mean}}, P_{\text{std}}) = (1, 1)$ as recommended by~\cite{chen2023importance}. We set the sequence length $T=25$ and context length $S=5$. For both phases, we use the AdamW optimizer with a learning rate of $10^{-5}$ and a batch size of 16. The training required approximately six days on 16 A6000 GPUs for the first phase and four days on 8 A100 GPUs for the second phase.

\subsection{Inference Details on In-the-Wild Videos}

For in-the-wild input videos, we first employ MegaSaM~\cite{li2024megasam} to estimate the camera intrinsics and poses, followed by generating the corresponding masked equirectangular video used to condition the network. After generation, we apply video super-resolution model~\cite{he2024venhancer} enhanced by our proposed blended decoding to increase the spatial resolution of the generated video by a factor of 2. Note that we do not apply super resolution modules in ablation studies and comparison with baseline methods.

\subsection{Metrics}

We evaluate our results based on three key criteria: image quality, temporal coherency, and geometric consistency. For image quality, we use PSNR, LPIPS~\cite{lpips}, Imaging Quality, and Aesthetic Quality metrics from VBench~\cite{vbench}. For temporal coherency, we employ FVD~\cite{unterthiner2019fvd} and the Motion Smoothness~\cite{vbench}. For geometric consistency, we introduce a \emph{line consistency} metric to evaluate whether straight lines remain straight within extrapolated views. This metric is particularly important for assessing whether our model preserves fundamental geometric properties when generating novel views. To quantitatively measure this consistency, we follow~\cite{qian2023understanding} and use EA-score~\cite{zhao2021deephough} to evaluate the angular and Euclidean distances between line pairs.

Specifically, FVD is calculated on the full 360$^\circ$ scene to evaluate overall distribution, while VBench metrics are applied to four square 2D projections (front, back, left, right) extracted from the 360$^\circ$ video, as VBench is designed for perspective videos. PSNR and LPIPS are computed only within masked regions of visible directions and aggregated across frames, since other directions are extrapolated. Though this visible region remains under-constrained (visible areas at timestamp $0$ may not appear at timestamp $T$), this approach provides more accurate evaluation than existing video outpainting methods~\cite{chen2024follow, dehan2022complete, be-your-outpainter} that calculate scores over the entire generated video.

\myparagraph{Line Consistency.} We introduce a line consistency metric to evaluate geometric fidelity across extrapolated viewpoints. This metric assesses whether straight lines in the original perspective remain consistent in neighboring views. Our approach uses real-world perspective videos that contain prominent linear structures, such as lanes and sidewalks. 

\begin{figure}[t!]
    \centering
    \includegraphics[width=\linewidth]{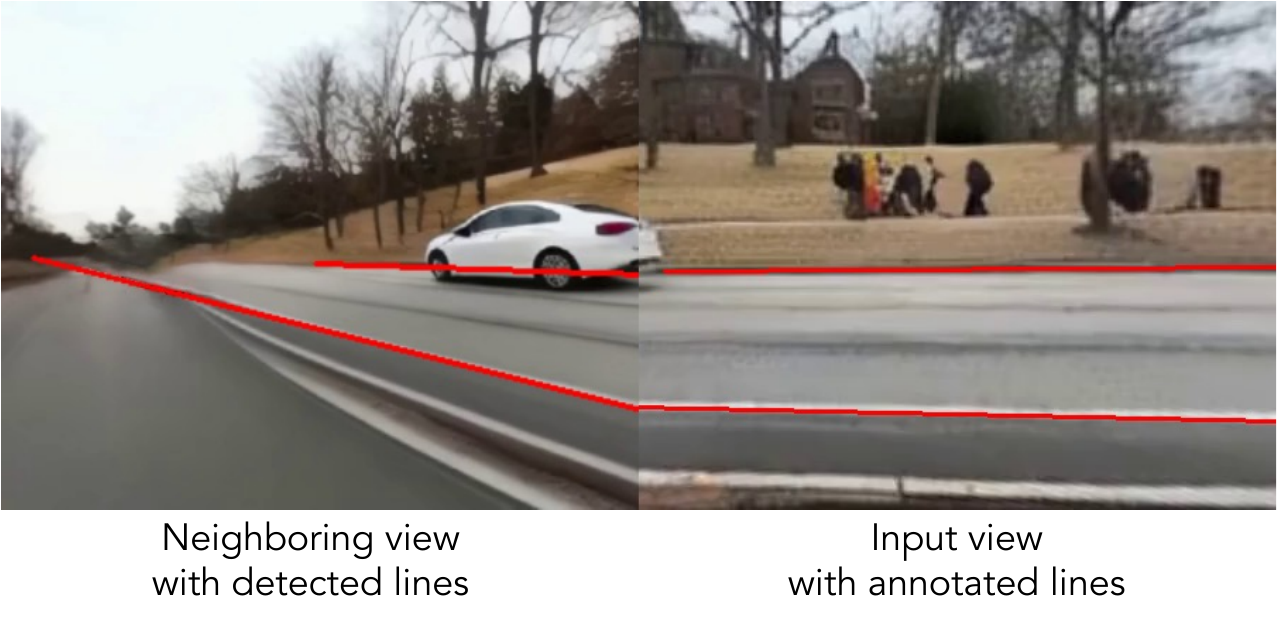}
    \vspace{-8mm}
    \caption{\textbf{Illustration of our line detection metric.} Given input view with annotated linear structures, we detect their extension in the neighboring views and measure their consistency.}
    \vspace{-3mm}
    \label{fig:line-detection}
\end{figure}

Specifically, we first annotate lines in input views, then detect corresponding lines in neighboring views unwrapped from generated 360° videos using the Hough transform. Then, we compute the analytical solution of ground truth lines in neighboring views using homography and employ bipartite matching to pair these with detected lines. Finally, we follow~\cite{qian2023understanding}~and report the EA-score~\cite{zhao2021deephough}, a score in $[0, 1]$ to measure the angle and euclidean distance between two lines, between the matched ground truth and detected lines. An example of our dataset and the line detection result in shown in Fig.~\ref{fig:line-detection}.

\begin{figure*}[t]
    \centering
    \includegraphics[width=\linewidth]{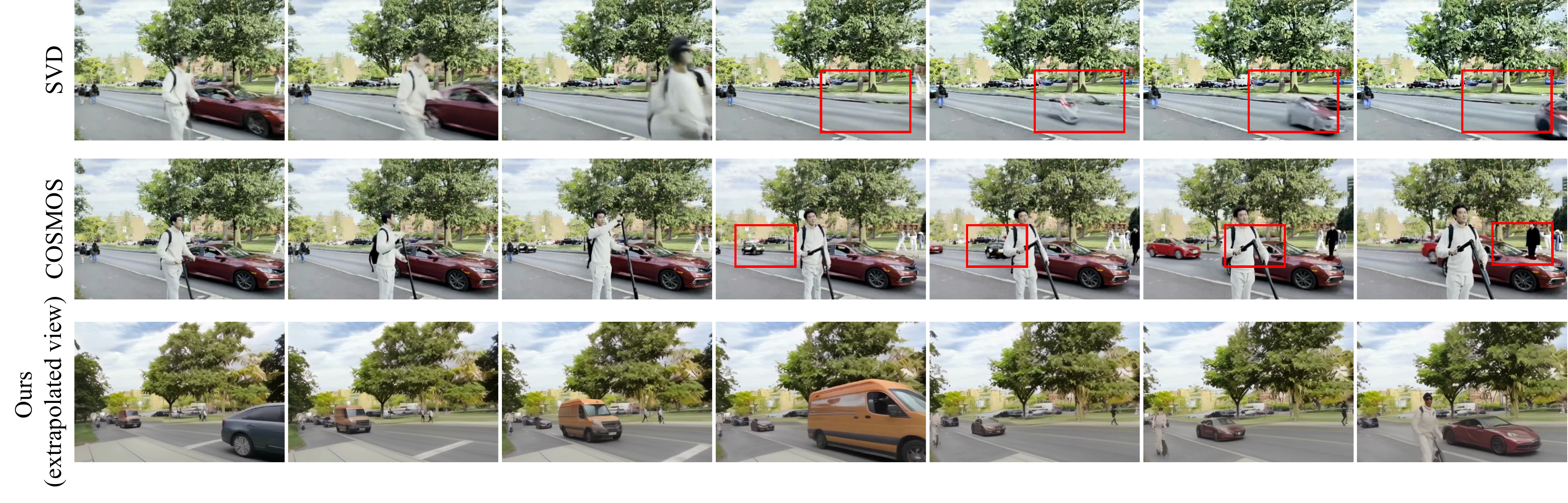}
    \vspace{-5mm}
    \caption{\textbf{Comparison with perspective video generation models.} Preserving shape consistency and dynamic plausibility remains an open challenge for video generation models. Specifically, our base model, SVD, exhibits noticeable appearance changes in the generated video (first row), while even state-of-the-art video models such as COSMOS demonstrate physical artifacts, where the black car on the back disappears (middle row).}
    \label{fig:svd-cosmos}
\end{figure*}

\subsection{Baseline Implementation Details}

\myparagraph{PanoDiffusion~\cite{wu2023panodiffusion}.} We reproduced this model due to the unavailability of their training code. We finetuned the image inpainting model~\cite{inpainting} on the video frames of our dataset, omitting the depth branch due to the lack of depth information in the dataset. The model was trained for 50K iterations using the AdamW optimizer with a learning rate of $10^{-5}$ and a batch size of 128, running on 8 NVIDIA A6000 GPUs.

\myparagraph{Be-Your-Outpainter~\cite{be-your-outpainter} and Follow-Your-Canvas~\cite{chen2024follow}.} Video outpainting methods support only rectangular inputs, so we centered square videos on the canvas and expanded the vertical field of view to 180$^\circ$ and horizontal field of view 360$^\circ$. For evaluation, we extracted three perspective videos from each 360$^\circ$ test video with FoVs of 60$^\circ$, 90$^\circ$, and 120$^\circ$. Because these models require per-video optimization for each generation, they are very compute expensive, taking about 14 and 11 minutes, respectively, on a single NVIDIA A6000 GPU for each generation. In contrast, our method does not introduce additional compute overhead upon SVD, taking around 90 seconds for each generation while achieving significantly better quality.

\myparagraph{Limitations.} Due to computational resource constraints, our current output resolution ($512\times 1024$) is lower than that of typical 4K real-world panoramas. The resolution further decreases when unwrapping back to perspective views. Additionally, while our model substantially improves upon the base SVD model in terms of object dynamics and temporal consistency, it still exhibits shape inconsistencies and physics artifacts, similar to SVD and other SoTA video models such as COSMOS, as shown in Figure~\ref{fig:svd-cosmos}.

\section{Additional Qualitative Results}

Additional comparison, application, and in-the-wild video generation results are available in our~\projectpagehref.